\documentclass[pmlr,twocolumn,10pt]{jmlr} 

\usepackage{booktabs}
\usepackage{siunitx}

\usepackage[switch]{lineno}

\usepackage{bm}
\usepackage{multirow}
\usepackage{multicol}
\usepackage{makecell}
\usepackage{lipsum}

\newcommand{\equal}[1]{{\hypersetup{linkcolor=black}\thanks{#1}}}

\newcommand{\ours}{LabTOP}
\newcommand{\oursfull}{Lab Test Outcome Predictor}

\theorembodyfont{\upshape}
\theoremheaderfont{\scshape}
\theorempostheader{:}
\theoremsep{\newline}

\jmlrvolume{287}
\jmlryear{2025}
\jmlrsubmitted{} 
\jmlrpublished{} 
\jmlrworkshop{Conference on Health, Inference, and Learning (CHIL) 2025}

\title[\ours{}]{LabTOP: A Unified Model for Lab Test Outcome Prediction\titlebreak
on Electronic Health Records}

\author{%
 \Name{Sujeong Im}\equal{These authors contributed equally} \Email{sujeongim@kaist.ac.kr}\\
 \Name{Jungwoo Oh}\footnotemark[1] \Email{ojw0123@kaist.ac.kr}\\
 \Name{Edward Choi} \Email{edwardchoi@kaist.ac.kr}\\
 \addr KAIST, Republic of Korea
}

\begin{document}

\maketitle

\begin{abstract}
Lab tests are fundamental for diagnosing diseases and monitoring patient conditions.
However, frequent testing can be burdensome for patients, and test results may not always be immediately available.
To address these challenges, we propose \textbf{\oursfull{} (\ours)}, a unified model that predicts lab test outcomes by leveraging a language modeling approach on EHR data.
Unlike conventional methods that estimate only a subset of lab tests or classify discrete value ranges, LabTOP performs continuous numerical predictions for a diverse range of lab items.
We evaluate LabTOP on three publicly available EHR datasets and demonstrate that it outperforms existing methods, including traditional machine learning models and state-of-the-art large language models.
We also conduct extensive ablation studies to confirm the effectiveness of our design choices.
We believe that LabTOP will serve as an accurate and generalizable framework for lab test outcome prediction, with potential applications in clinical decision support and early detection of critical conditions.

\end{abstract}
\paragraph*{Data and Code Availability}
This paper uses the three EHR datasets, MIMIC-IV~\citep{johnson2023mimic}, eICU~\citep{pollard2018eicu}, and HiRID~\citep{hyland2020early}, which are publicly available on the PhysioNet repository~\citep{johnson2024physionet, pollard2019physionet, yeche2021physionet}.
More details about datasets can be found at Section~\ref{sec:data}.
Our implementation code can be accessed at this repository.\footnote{\url{https://github.com/sujeongim/LabTOP}}

\paragraph*{Institutional Review Board (IRB)}
This research does not require IRB approval.

\section{Introduction}
\label{sec:intro}

Electronic Health Records (EHR) are essential to modern healthcare systems, serving as comprehensive databases of patient data, including treatments, clinical interventions, and lab test results~\citep{gunter2005emergence}.
These records provide a longitudinal view of a patient's medical history, allowing for the tracking of individual health trends~\citep{kruse2017impact}.
Among them, lab test results play a particularly important role by capturing numerical changes in key biomarkers, such as blood glucose and creatinine.
These results reflect a patient’s physiological and pathological state, supporting the management of disease and the assessment of treatment efficacy~\citep{sikaris2017enhancing, cabalar2024role}.

Despite their clinical importance, conducting lab tests often faces challenges in real-world settings.
For instance, the need for frequent lab tests in unstable patients conflicts with the increased distress they experience from repeated invasive procedures, creating a trade-off between clinical necessity and patient burden~\citep{ambasta2019expert,zhi2024hgbnet}.
Additionally, some tests often take significant time to obtain results, making it difficult to assess the patient's condition promptly~\citep{shiferaw2019magnitude}.
As a result, there are limitations in comprehensively assessing a patient's condition from multiple perspectives through various lab tests, forcing healthcare providers to rely solely on the available test results to make clinical judgments.
This challenge highlights the need for methods to estimate lab test results without conducting the actual tests, allowing healthcare providers to understand a patient's condition comprehensively through the predicted lab test results.

Meanwhile, the advancement of machine learning has greatly contributed to healthcare for several clinical prediction tasks such as readmission, mortality risk, and length of hospital stay~\citep{song2018attend, xiao2018readmission,hur2022unifying,hur2023genhpf,kim2023general,wornow2023ehrshot, renc2024zero}.
While previous studies have demonstrated the value of lab test data in supporting clinical decisions, efforts to estimate various lab test results within a single unified model have been limited.
Notably, most prior studies approach lab test outcome prediction by classifying discrete levels of a small subset of selected lab items~\citep{hur2023genhpf,kim2023general,wornow2023ehrshot}.
Even among studies that attempt to estimate continuous values, they are specifically designed for only a few selected lab tests rather than whole set of lab items~\citep{zhi2024hgbnet, jiang2024probabilistic, duan2020ngboost, liu2023machine, fu2023implementation, langarica2024deep}.
On the other hand, some works on irregularly sampled time-series models have explored the prediction of multiple lab measurements using neural ODE-based approaches~\citep{rubanova2019latent, schirmer2022modeling}.
These methods estimate multiple lab values within a unified framework; however, they primarily focus on modeling the dynamics of lab test results alone, without incorporating the broader medical events present in EHRs, such as medications and input events.
This narrow focus limits their practical applicability in clinical settings, where precise numerical predictions for a broader range of lab test results can facilitate early intervention and support more informed clinical decision-making across diverse medical conditions.

To address these limitations, we propose \textbf{\oursfull{} (\ours{})}, a novel method for predicting a diverse range of lab test outcomes based on a patient's medical history within a single unified model.
Inspired by autoregressive models widely used in Natural Language Processing~\citep{radford2018improving, radford2019language}, we leverage language modeling paradigm to predict numeric values in an autoregressive manner, instead of regressing each lab test value separately.
This approach enables us to build a unified model capable of predicting outcomes for multiple lab items within a single framework, without requiring us to train a separate model for each lab item.
Furthermore, unlike prior approaches, \ours{} is designed to estimate continuous lab test measurements across a wide spectrum of biomarkers, offering greater granularity and broader applicability in clinical practice.
The main contributions of this work can be summarized as follows:
\begin{itemize}
    \item We propose \ours{}, a new model that estimates continuous values for a wide range of lab test measurements given the patient's previous medical history.
    To the best of our knowledge, this is the first comprehensive approach to predict various lab test results within a single unified model trained on EHR data.

    \item \ours{} outperforms existing approaches, including a recent large language model (LLM), across three publicly available datasets.
    This demonstrates the generalizability of our method to diverse patient populations and clinical contexts.

    \item Considering the time-series characteristics of EHR data where each patient can be represented as a sequence of medical events, we conduct ablation studies to explore the best way to represent the time and numeric features of medical events for the model.
    We believe these experiments will provide valuable insights into effectively modeling EHR data.

\end{itemize}

\section{Related Works}

\paragraph{Predicting clinical outcomes using EHR data}
Previous machine learning models leveraging EHR data have primarily focused on predicting clinical outcomes, such as readmission, mortality risk, and length of stay~\citep{song2018attend, xiao2018readmission,hur2022unifying,hur2023genhpf,kim2023general,wornow2023ehrshot, renc2024zero}.
Among them, \citet{song2018attend} introduced the \textit{SAnD}, employing Transformer architecture~\citep{vaswani2017attention} to predict in-hospital mortality and length of hospital stay.
They constructed the input embeddings based on structured clinical codes (e.g., ICD, LOINC, RxNorm) to process EHR data, which makes it difficult to process multiple EHRs with different schemas within a single model.
To address these challenges, \citet{hur2022unifying} proposed a text-based embedding approach (DescEmb) to represent EHR events using descriptive text rather than codes, showing superior performances compared to the code-based approach.
Based on this text-based approach, GenHPF~\citep{hur2023genhpf} extended it by incorporating multi-task learning to handle various predictive tasks simultaneously.
Furthermore, REMed~\citep{kim2023general} integrated an event-retriever module into this framework to selectively extract relevant events with target tasks, ensuring that only the important events contribute to the final predictions.

\paragraph{Generative models for EHR data}
In addition to the studies employing discriminative models for predictive tasks, generative approaches have also been explored for modeling EHR data.
Specifically, MedGPT~\citep{kraljevic2021medgpt} adopted the GPT-2 architecture~\citep{radford2019language} to predict next SNOMED-CT disorder concept given the patient's previous history of disorders.
Although this work utilized only the disorder concepts instead of a comprehensive set of clinical events in EHR, they demonstrated the potential of generative methods for simulating patient trajectories and predicting future clinical events.
In addition, ETHOS~\citep{renc2024zero} employed a generative approach based on language modeling to simulate clinical scenarios using a structured EHR data.
Similar to the code-based approach, they converted each medical event into 1 to 7 tokens to construct the input sequence for the Transformer~\citep{vaswani2017attention}, and trained the model to predict every next token given the prior sequences to simulate the clinical outcomes of the patient, such as whether the patient will die or not (\textit{i.e.,} mortality prediction).

\paragraph{Lab test outcome prediction}
Although previous studies have significantly advanced the field of machine learning for healthcare, only a few of them have addressed the direct estimation of continuous values for various lab tests within a single unified model.
For example, there have been attempts to estimate multiple lab values within a unified framework using neural ODE-based approaches~\citep{rubanova2019latent,schirmer2022modeling}.
However, a key limitation of these approaches is that they only model lab test values in isolation, without incorporating other essential patient events such as medications, interventions, and clinical inputs that may significantly influence lab test dynamics.
Instead, most of previous studies have explored the prediction of lab test results by designing a specific model for only a few selected lab items or classifying discrete levels of selected lab items, such as normal or abnormal ranges.
For example, HgbNet~\citep{zhi2024hgbnet} employed an attention-based LSTM model to estimate hemoglobin levels, while \citet{jiang2024probabilistic} presented a probabilistic model using NGBoost~\citep{duan2020ngboost} to predict expected values of several lab items, such as wbc and hemoglobin.
Similarly, several works~\citep{liu2023machine,fu2023implementation,langarica2024deep} attempted to estimate glucose values using machine learning or deep learning approach.
However, these approaches are specifically designed for a small subset of lab tests, making it challenging to generalize across diverse clinical measurements.

Meanwhile, GenHPF~\citep{hur2023genhpf} and REMed~\citep{kim2023general} formulated this task as classifying the discrete level of values for several selected lab items.
Although these works have shown high performances on the defined task, their predictions remain limited in terms of extensibility and granularity.
First, they rely on predefined level thresholds for each lab item, determined through domain knowledge, which makes it difficult to extend the approach to encompass all lab items recorded in the EHR database.
Second, as the task is defined within a fixed prediction window (\textit{e.g.,} predicting average levels of lab items in the next 12 hours given the first 12 hours), these models lack granularity in providing immediate estimations of lab test results, which is critical for clinical scenarios where timely and dynamic predictions are necessary to inform urgent decisions.

\begin{figure*}[ht] 
\floatconts
    {fig:overview}
    {\vspace{-8mm}
    \caption{\textbf{Training and Inference of \ours{}.} During training, demographic information $D$ and a sequence of medical events $P=[\mathcal{M}_1,\dotsc,\mathcal{M}_N]$ are tokenized, and fed into the Transformer to generate next token at each position. The training loss is then computed only for lab test value tokens along with their units and the corresponding $\texttt{[EOE]}$ token. During inference, given the preceding sequence of medical events and the target lab event (\textit{i.e.,} $(t_k,e_k,d_k)$), the model autoregressively generates numeric value tokens for its outcome until the [EOE] token is encountered.}}
    {
        \includegraphics[width=1.0\textwidth]{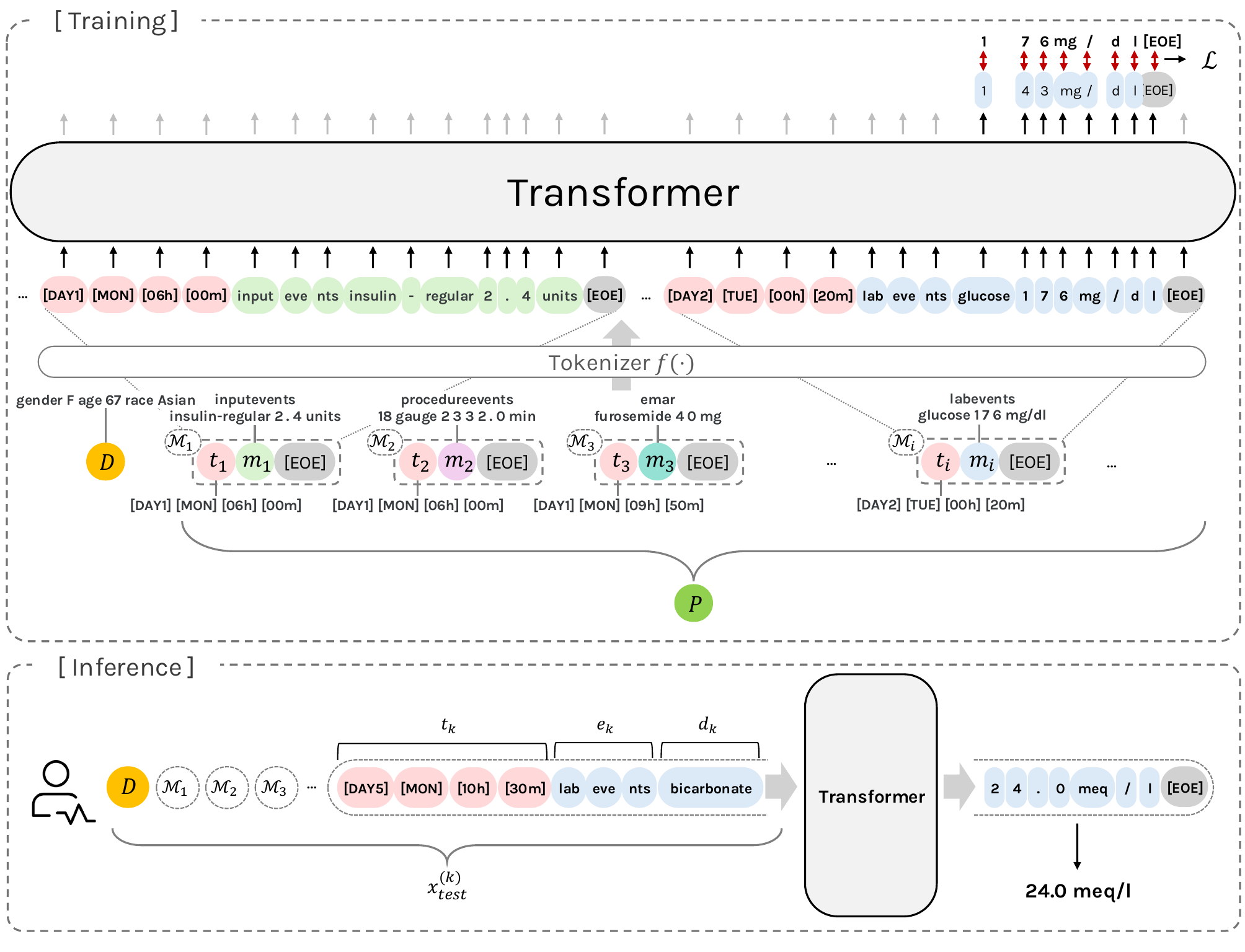}
    }
    \vspace{-6mm}
\end{figure*}

\section{Methods}

\subsection{Preliminaries}
In EHR data, patient medical events that occur in a hospital, such as lab tests and prescriptions, are recorded in a structured format.
Thus, we can represent a patient $P$ as a sequence of medical events $[\mathcal{M}_1, \dotsc, \mathcal{M}_N]$, where $N$ is the total number of medical events, and $\mathcal{M}_i$ is the $i$-th medical event for the patient.  
Each medical event typically includes a timestamp, medical event codes (\textit{e.g.,} ICD, LOINC, RxNorm), a numerical value, and its corresponding unit of measurement.
Accordingly, the $i$-th medical event $\mathcal{M}_i$ can be represented as a 5-tuple of event features $(t_i, e_i, c_i, v_i, u_i)$, where $t_i$ represents the timestamp of the event, $e_i$ stands for the type of the event (\textit{e.g.,} ``labevent'', ``prescription''), $c_i$ is the corresponding medical code (\textit{e.g.,``2160-0"}), $v_i$ denotes the measured values (\textit{e.g.,} 1.2), and $u_i$ is the associated unit of measurement (\textit{e.g., mg/dL}).
The detailed definitions of these event features are described in the following section, as well as depicted in Figure~\ref{fig:overview}.

\subsection{Preprocessing}

\paragraph{Textual representation of medical events}
Unlike conventional methods that directly embed medical code $c_i$ into a fixed-dimensional vector space to represent a medical event $\mathcal{M}_i$~\citep{miotto2016deep, choi2016doctor, choi2020learning, song2018attend, shang2019pre, renc2024zero}, we start by converting each medical code $c_i$ (\textit{e.g.,``2160-0"}) into its corresponding textual description $d_i$ (\textit{e.g.,``Creatinine [Mass/volume] in Serum or Plasma"}).
Additionally, we also treat its numerical value $v_i$ as plain text and split it into each digit, so that each digit is processed as a separate token in the input sequence~\citep{hur2022unifying, hur2023genhpf, kim2023general}.
In this way, we reconstruct a 4-subtuple of $(e_i, c_i, v_i, u_i)$ included in the medical event $\mathcal{M}_i$ into a textual sequence, and denote as $m_i$. 
For example, a lab test event standing for a creatinine value of 1.23 mg/dL measured at the time of $t_i$ is transformed into $(t_i, m_i)$ where $m_i$ is a plain text of ``labevent creatinine 1~ .~ 2~ 3 mg/dL''. 

\paragraph{Absolute time encoding}
\label{sec:abs_time_encoding}
Temporal information also plays a crucial role in modeling the progression of a patient's health status accurately, as the timing and frequency of the events provide essential insights into clinical trajectories.
While previous studies have primarily employed a relative time encoding strategy to represent temporal intervals between consecutive events~\citep{hur2023genhpf, renc2024zero}, this approach may not always be optimal due to the following reasons.
Medical events for inpatients, such as prescriptions and lab tests, usually follow regular patterns dictated by hospital routines, clinical protocols, or daily schedules.
Additionally, a patient's physiological status can also depend on time-of-day patterns, making absolute time representation more effective in capturing meaningful trends.
For example, blood glucose measurements are heavily affected by a patient's fasting state~\citep{moebus2011impact}.
By incorporating absolute time encoding, the model can infer contextual factors such as feeding schedules or overnight fasting periods, leading to improved predictive performances for lab test outcomes.

To achieve this, we represent each timestamp $t_i$ as a combination of four types of special tokens, capturing essential temporal attributes that influence patient health status.
Specifically, these tokens include (1) the elapsed days since the ICU admission, (2) the day of the week the event occurred, (3) the absolute hour of the event, and (4) the aggregated minute value.
The minute value is aggregated in 10-minute intervals to balance granularity and computational efficiency, ensuring that time representations remain meaningful without introducing excessive sparsity.
For example, for the event recorded on the first day since the ICU admission, at 13:38 on a Tuesday, the timestamp $t_i$ is defined as $[\texttt{[DAY1]},\texttt{[TUE]},\texttt{[13h]}, \texttt{[30m]}]$.
In addition, we append a special $\texttt{[EOE]}$ token indicating the end of the event to every textual medical event $m_i$ and concatenate a sequence of medical events, so that a patient $P$ is now represented as
\begin{equation}
    P = \Big\|_{i=1}^{N}[t_i, m_i, \texttt{[EOE]}]
\end{equation}
where $\|$ is a symbol denoting the sequence concatenation operator.

\paragraph{Demographic encoding}
Physiological ranges for lab tests can vary significantly based on demographic attributes such as age, gender, and race.
For example, creatinine levels tend to be higher in males due to greater muscle mass~\citep{culleton1999prevalence}, while hemoglobin levels may vary depending on both gender and age~\citep{murphy2014sex}.
To incorporate this information into our model, we convert each patient's demographic attributes into a textual representation, denoted as $D$.
For example, for a 65-year-old Asian female patient, $D$ is defined as ``gender F age 65 race Asian''.
We prepend this demographic information to the patient's sequence of medical events $P$ and tokenize the sequence using the Bio-Clinical BERT tokenizer~\citep{alsentzer2019publicly} to obtain the final input sequence for the model, which is defined as
\begin{equation}
    x = f(D \| P) = f(\Big[D, \Big\|_{i=1}^{N}[t_i, m_i, \texttt{[EOE]}] \Big])
\end{equation}
 where $f(\cdot)$ denotes the tokenization function that converts the text sequence into a sequence of token indices.

\subsection{LabTOP Training}
\paragraph{Model architecture}
We design our model following the GPT-2 architecture~\citep{radford2019language}, which is composed of Transformer decoder layers~\citep{vaswani2017attention}.
Specifically, the input sequence $x$ is embedded into a sequence of latent vectors $\bm{X}\in\mathbb{R}^{M \times h}$ by the learnable embedding layer, where $M$ is the length of the input token sequence and $h$ is the hidden dimension of the model.
They are then fed into the Transformer decoder to model the conditional probability of the next token given the previous sequence, following an autoregressive language modeling approach.

\paragraph{Objective}
Unlike traditional language modeling where the objective is to predict the next token for every position in the sequence, we modify the loss computation to focus exclusively on lab test outcome prediction.
That is, instead of computing the loss for all tokens, we selectively compute the loss only on the tokens corresponding to lab test values, encouraging the model to focus on accurately predicting lab test outcomes while still benefiting from a broader contextual understanding of the patient's medical history.
Accordingly, the loss is defined as
\begin{equation}
    \mathcal{L}=- \sum_{i\in{\mathcal{I}_{lab}}}\log{p(x_i|x_{<i};\theta)}
\end{equation}
where $\mathcal{I}_{lab}$ represents the set of token positions corresponding to lab test values along with their unit of measurements, as well as their special $\texttt{[EOE]}$ token.

\paragraph{Random permutation}
As mentioned in the previous section, timestamps are discretized into 10-minute interval tokens (\textit{e.g.,} 13:38 $\rightarrow$ $[\texttt{[13h]},\texttt{[30m]}]$), which can result in multiple medical events having the same timestamp tokens.
Accordingly, in order to make the model invariant to their ordering during training, we apply random permutation to medical events that share the same timestamp.
Specifically, given a set of events having the same timestamp, we randomly shuffle the order of the events at each training iteration to prevent the model from overfitting to any arbitrary order within the same timestamp.

\subsection{LabTOP Inference}
We formulate the lab test outcome prediction task as a language modeling task to predict numeric values token-by-token in an autoregressive manner.
That is, the model generates the corresponding numeric value of a target lab item given the sequence of prior medical events and the target lab item's name.
Accordingly, to construct inference samples, we extract every lab item appearing in the input sequence and build a corresponding inference prompt.
Specifically, for each target lab test item occurring at timestamp $t_k$, we concatenate all prior medical events up to $t_k$ along with the target lab test's textual representation, $e_k$ and $d_k$, to create an inference sequence.
Thus, the $k$-th inference sample of a patient is defined as
\begin{equation}
    x_{test}^{(k)} = f(\Big[D, \Big\|_{t_i<t_k}[t_i, m_i, \texttt{[EOE]}],[t_k,e_k,d_k]\Big])
\end{equation}
where $e_k$ and $d_k$ denote the event type (\textit{i.e.,} ``labevent'') and textual description of the target lab test (\textit{e.g.,} ``bicarbonate''), as aforementioned.

Once the inference sequence is constructed, it is fed into the trained model, which generates tokens autoregressively until the $\texttt{[EOE]}$ token is encountered.
In this process, our primary objective is to accurately predict the outcome value of a specific lab test at a given timestamp rather than probabilistically generating potential future lab tests.
Thus, to ensure precise predictions, we do not sample from the model's output distribution but instead select the most probable token (\textit{i.e.,} top-1 token) at each decoding step.
The generated sequence is then decoded back into a numerical value (\textit{e.g.,} 2 4 . 0 meq / l $\rightarrow$ 24.0 meq/l).

\section{Experiments}
\subsection{Data}
\label{sec:data}
To demonstrate the generalizability of our approach, we conduct experiments on three publicly available EHR datasets: MIMIC-IV~\citep{johnson2023mimic}, eICU~\citep{pollard2018eicu}, and HiRID~\citep{yeche2021hirid}.
Specifically, MIMIC-IV consists of 94,458 ICU records of 364,627 patients from the Beth Israel Deaconess Medical Center, including structured time-series data such as lab test results, and medication administrations.
eICU is a large-scale, multi-center dataset that includes 200,859 ICU records of 139,367 patients from hospitals across the United States, consisting of a broader range of patient demographics and treatment variations compared to single-center datasets.
HiRID is a high-resolution ICU dataset collected from the Bern University Hospital in Switzerland, composed of 33,905 ICU patients.
For all these datasets, we take the ICU records whose length of stay is at least 6 hours.
We then randomly split them into training, validation, and test sets in an 8:1:1 based on their patient IDs, and treat each ICU record as an individual sample for our model.
Detailed dataset statistics are described in Appendix~\ref{app:data_statistics}.

To determine which types of EHR events should be involved for lab test outcome prediction, we categorized EHR events into two broad types: observation-type events, which represent the patient's physiological state (\textit{e.g.,} lab tests, microbiology tests), and intervention-type events, which correspond to medical treatments, prescriptions, and procedures administered during the ICU stay.
We then prioritized these event types based on their expected influence on lab test results:
(1) Firstly, medication events that directly reflect the patient's physiological status by indicating administered drugs and dosages.
(2) Secondly, other intervention-related events, including input events (\textit{e.g.,} fluid intake) and procedure events, which affect the patient's overall condition.
(3) Lastly, observation-type events that do not directly alter a patient's state but provide useful contextual information about their physiological trends.
Based on these criteria, the following tables for each dataset are used:
\begin{itemize}
    \item MIMIC-IV: \textit{labevents}, \textit{emar}, \textit{emar\_detail}, \textit{inputevents}, \textit{procedureevents}, \textit{outputevents}, \textit{microbiologyevents}

    \item eICU: \textit{lab}, \textit{medication}, \textit{infusiondrug}, \textit{treatment}, \textit{intakeoutput}, \textit{microlab}

    \item HiRID: \textit{observation\_tables}, \textit{pharma\_records}
\end{itemize}
Note that HiRID collected every observation-type event into one table (\textit{i.e.,} \textit{observation\_tables}), which leads to multiple types of events, such as lab tests and microbiology tests, being contained in this single table.
However, there is only an indicator specifying whether an event corresponds to a lab test or not, so we cannot differentiate other observation-type events within this table.
Accordingly, we select only those events explicitly identified as lab tests, excluding all other events from this table.

Additionally, to ensure a focus on the most clinically relevant medical events, we select only the medical items that account for the top 90\% of occurrences within their respective tables for each dataset, including the lab test tables as well as other event types such as medication, input events, and procedure events.
As a result, we select 44 unique lab tests from MIMIC-IV, 41 from eICU, and 27 from HiRID.
The included lab items from each dataset are detailed in Appendix~\ref{app:data_statistics}.

\subsection{Evaluation Metrics}
\paragraph{NMAE}
Normalized Mean Absolute Error (NMAE) is a normalized version of Mean Absolute Error (MAE) that accounts for differences in scale across multiple variables with different units and ranges.
Unlike standard MAE, which directly measures the average absolute difference between the predicted and ground-truth values, NMAE adjusts for variations in measurement scales by normalizing each error term based on the range of the corresponding lab item.
Specifically, for each unique lab item $l$, we compute its scale using the difference between the 99th percentile and the 1st percentile values (\textit{i.e.,} $v_{99\%}^{(l)} - v_{1\%}^{(l)}$) in the test set to mitigate the influence of extreme outliers.
The NMAE for a given lab test $l$ is then computed by dividing its MAE by this scale, which is defined as
\begin{equation}
    \text{NMAE}_{l}=\frac{1}{v_{99\%}^{(l)}-v_{1\%}^{(l)}}\sum_{i\in\mathcal{I}_l}\frac{|y_i-\hat{y}_i|}{N_l}
\end{equation}
where $\mathcal{I}_l$ represents the set of sample indices corresponding to the target lab test, $y_i$ and $\hat{y}_i$ denote the ground-truth and predicted value for $i$-th sample, and $N_l$ is the number of target lab samples in the test set.

\paragraph{SMAPE}
Symmetric Mean Absolute Percentage Error (SMAPE) is a widely used metric to evaluate the accuracy of prediction for continuous values, providing an intuitive measure of how much predicted values deviate from ground-truth values in percentage terms.
Unlike traditional Mean Absolute Percentage Error (MAPE), SMAPE ensures a symmetric scaling by normalizing the error terms based on the sum of the absolute values of the ground-truth and predicted values, preventing over-penalization when the ground-truth values are small.
This property makes SMAPE especially suitable for evaluating lab test outcome predictions, where the ranges of different lab items can vary significantly.
The SMAPE for a given lab test $l$ is defined as
\begin{equation}
    \text{SMAPE}_l = \frac{1}{N_l} \sum_{i \in \mathcal{I}_l} \frac{|y_i - \hat{y}_i|}{(|y_i| + |\hat{y}_i|) / 2} \times 100
\end{equation}

\subsection{Baselines}
To demonstrate the effectiveness of \ours, we compare its performance with the following baseline methods, including both traditional machine learning approaches and state-of-the-art LLMs.
\begin{itemize}
    \item \textbf{Naive}: A simple method that predicts the lab test value at a given timestamp by using the most recent measured value of the same lab test, commonly referred to as the naive forecasting method.
    \item \textbf{Naive($\mu$)}: A variant of the naive method that predicts the lab test value by averaging all previously recorded values of the same lab test before the prediction timestamp.
    \item \textbf{GenHPF}~\citep{hur2023genhpf}: A multi-task learning model that incorporates hierarchical patient features for clinical prediction tasks. We adapt GenHPF to estimate lab test values by formulating it as a regression task.
    \item \textbf{XGBoost}~\citep{chen2016xgboost}: A widely used gradient boosting method designed for structured data. For this baseline, we use several statistics of past measurements, such as count and mean.
    \item \textbf{GPT-4o} and \textbf{GPT-4o-mini}: LLMs designed for general-purpose tasks. We evaluate these models to assess the capability of general-purpose models in predicting lab test outcomes. Since these models process long patient histories as input, running inference across the entire test set would require substantial token usage. To balance computational feasibility with representative evaluation, we sample 10\% of the test set\footnote{To be more specific, running inference in a batch on the sampled 10\% of the test set incurred a total cost of approximately \$500.}. All experiments employing GPTs were executed using the HIPAA-compliant GPT model available on Azure.\footnote{\url{https://learn.microsoft.com/en-us/azure/compliance/offerings/offering-hipaa-us}}
    \item \textbf{LLaMA-3.1-Instruct (8B)}~\citep{grattafiori2024llama}: We also evaluate the open-sourced instruction-tuned LLM designed for general-purpose tasks.
    For the experiments on this large model, we sample 10\% of the test set as the same with GPT-4o and GPT-4o-mini.
\end{itemize}
More details about each model's implementation, such as training hyperparameters and feature engineering for XGBoost, are provided in Appendix~\ref{app:implementation_details}.

\subsection{Results}

\begin{table}[t]
\floatconts
    {tab:main}
    {\caption{
    Lab test outcome prediction performances on different EHR datasets, where mean and 95\% confidence interval are shown across 3 different random seeds for trainable models (e.g., GenHPF, XGBoost, and LabTOP).
    The best performances for each dataset are highlighted with \textbf{boldface}.}
    \vspace{-5mm}
    }
    {\resizebox{1.0\linewidth}{!}{
        \begin{tabular}{lccc}
            \toprule
             & MIMIC-IV & eICU & HiRID \\
            \midrule
            \multicolumn{4}{l}{\textit{\textbf{NMAE}}} \\
            Naive & $0.090$ & $0.102$ & $0.101$ \\
            Naive($\mu$) & $0.108$ & $0.116$ & $0.119$ \\
            GenHPF & $0.112_{\pm 0.007}$ & $0.106_{\pm 0.011}$ & $0.127_{\pm 0.027}$\\
            XGBoost & $0.083_{\pm 0.000}$ & $0.475_{\pm 0.000}$& $0.100_{\pm 0.000}$ \\
            GPT-4o & $0.195$ & $0.230$ & $0.165$ \\
            GPT-4o-mini & $0.792$ & $1.027$ & $2.969$ \\
            LLaMA-3.1-Inst. & $1.271$ & $1.570$ & $3.280$ \\
            \noalign{\vskip 0.4ex}
            \hline
            \noalign{\vskip 0.4ex}
            \textbf{\ours{}} & $\bm{0.064_{\pm 0.001}}$ & $\bm{0.080_{\pm 0.007}}$ & $\bm{0.083_{\pm 0.004}}$ \\
            \midrule
            \multicolumn{4}{l}{\textit{\textbf{SMAPE (\%)}}} \\
            Naive & $17.64$ & $18.29$ & $\bm{15.98}$ \\
            Naive($\mu$) & $21.78$ & $20.83$ & $18.99$ \\
            GenHPF & $25.58_{\pm 0.52}$ & $24.22_{\pm 1.30}$ & $19.18_{\pm 2.98}$\\
            XGBoost & $20.42_{\pm 0.00}$ & $45.86_{\pm 0.00}$ & $17.18_{\pm 0.00}$ \\
            GPT-4o & $19.67$ & $21.18$ & $21.17$ \\
            GPT-4o-mini & $40.86$ & $45.67$ & $43.88$ \\
            LLaMA-3.1-Inst. & $75.58$ & $79.92$ & $76.63$ \\
            \midrule
            \textbf{\ours{}} & $\bm{14.80_{\pm 0.13}}$ & $\bm{15.96_{\pm 0.22}}$ & $16.83_{\pm 0.21}$ \\
            \bottomrule
        \end{tabular}
    }}
\vspace{-4mm}
\end{table}
The average results of lab test outcome prediction for three datasets are presented in Table~\ref{tab:main}.

\ours{} achieves the best performance in both NMAE and SMAPE across all datasets, except for NMAE in HiRID, where it still performs competitively.
Notably, naive approaches (Naive and Naive($\mu$)) show relatively higher SMAPE in HiRID compared to MIMIC-IV and eICU.
We speculate that this is due to the characteristics of HiRID, where lab tests are conducted more frequently than in the other two datasets.
Specifically, we found that the average interval between lab measurements is about 15.8 hours in MIMIC-IV, 19.8 hours in eICU, and only 12.9 hours in HiRID.
This shorter interval likely benefits methods that rely on simple statistical heuristics, such as carrying forward the last measured value (Naive) or computing historical averages (Naive($\mu$)), making them more effective in this setting.
However, our primary motivation is to predict lab test outcomes in environments where frequent testing is not always feasible.
Given this perspective, the strong performance of \ours{} on MIMIC-IV and eICU highlights its effectiveness in scenarios where repeated testing may not be feasible.

Interestingly, \ours{} also outperforms general-purpose LLMs (GPT-4o, GPT-4o-mini, and LLaMA-3.1-Instruct), confirming that these models struggle with lab test outcome prediction task.
This underscores the importance of explicitly learning EHR data rather than relying on general LLM capabilities.
Furthermore, GPT-4o-mini and LLaMA-3.1-Instruct exhibit significantly lower performance, which we attribute to its tendency to generate unrealistic lab test values.
Upon manual inspection, we found that GPT-4o-mini and LLaMA-3.1-Instruct often output numerically implausible values that are far outside the expected range (\textit{e.g.,} predicting \textit{ph of arterial blood} as 135 whereas the ground-truth is 7.2).
This result demonstrates that our data processing and training strategy play a crucial role in achieving robust performance by effectively leading the model to produce clinically meaningful predictions.

\subsection{Ablation Studies}
To systematically evaluate the contributions of different components in our model, we conduct ablation studies along the following perspectives.
All experiments in this section are conducted using the MIMIC-IV dataset to ensure consistency.

\begin{table}[t]
\floatconts
    {tab:abl_unified}
    {\caption{Ablation study results from different perspectives on MIMIC-IV dataset.} \vspace{-5mm}}
    {\resizebox{1.0\linewidth}{!}{
        \begin{tabular}{lccc}
            \toprule
            Strategy & & NMAE & SMAPE (\%) \\
            \midrule
            \multicolumn{2}{l}{\textit{Embedding \& training strategy}} & & \\
            \multicolumn{2}{l}{code-based \& Full AR} & $0.106_\pm{0.005}$ & $20.97_\pm{0.33}$ \\
            \multicolumn{2}{l}{code-based \& Only Lab AR} & $0.098_\pm{0.002}$ & $19.44_\pm{0.55}$ \\
            \multicolumn{2}{l}{\multirow{3}{*}{\makecell[l]{text-based \& Only Lab AR\\\textbf{(\ours)}}}} & \multirow{3}{*}{$\bm{0.064_{\pm 0.001}}$} & \multirow{3}{*}{$\bm{14.80_{\pm 0.13}}$} \\
            \\
            \\
            \hline
            \noalign{\vskip 0.4ex}
            \multicolumn{2}{l}{\textit{Numeric value representation}} & & \\
            \multirow{3}{*}{\makecell[l]{quantile-based\\}} & 5-quantiles & $0.077_\pm{0.001}$ & $18.27_\pm{0.11}$ \\
            & 10-quantiles & $0.072_\pm{0.000}$ & $16.64_\pm{0.14}$\\
            & 20-quantiles &$0.067_\pm{0.001}$ & $15.65_\pm{0.01}$ \\
            \multicolumn{2}{l}{\multirow{3}{*}{\makecell[l]{digit-wise tokenization\\\textbf{(\ours)}}}} & \multirow{3}{*}{$\bm{0.064_{\pm 0.001}}$} & \multirow{3}{*}{$\bm{14.80_{\pm 0.13}}$} \\
            \\
            \\
            \hline
            \noalign{\vskip 0.4ex}
            \multicolumn{2}{l}{\textit{Timestamp representation}} \\
            \multicolumn{2}{l}{relative time encoding} & $0.134_\pm{0.001}$ & $29.02_\pm{0.05}$ \\
            \multicolumn{2}{l}{\multirow{3}{*}{\makecell[l]{absolute time encoding\\\textbf{(\ours)}}}} & \multirow{3}{*}{$\bm{0.064_{\pm 0.001}}$} & \multirow{3}{*}{$\bm{14.80_{\pm 0.13}}$} \\
            \\
            \\
            \hline
        \end{tabular}
    }}
\vspace{-2mm}
\end{table}
\subsubsection{Embedding and Training Strategy}
EHR events can be processed using different embedding strategies, which we categorize into \textbf{code-based} and \textbf{text-based} representations.
The code-based approach directly embeds structured medical codes (\textit{e.g.,} LOINC, ICD), while the text-based approach converts medical codes into their textual descriptions before embedding.
Additionally, unlike conventional methods that apply autoregressive loss to all tokens (\textit{i.e.,} \textbf{Full AR}), we specifically designed our model to focus on lab test values by applying autoregressive loss selectively (\textit{i.e.,} \textbf{Only LAB AR}).
To examine the impact of these design choices, we compare our model trained with different combinations of embeddings and loss strategies.

In the code-based embedding strategy, a new vocabulary is defined to accomodate structured medical tokens, following the embedding approach introduced in ETHOS~\citep{renc2024zero}.
Specifically, one or two tokens are defined per event: one token for the total event features that encodes the event type, item name and unit of measurement (if applicable) as a single entity (\textit{e.g.,} [LAB\_inr(pt)] or [LAB\_glucose\_mg/dL]), and the other token for the corresponding value mapped to one of 10 quantile tokens if the event has a numeric value.
Additionally, for medications, we map drug names to their corresponding ATC (Anatomical Therapeutic Chemical) codes whenever available.

As shown in Table~\ref{tab:abl_unified}, the incremental application of our embedding and training strategy components consistently improves performance.
Specifically, switching from training all tokens in an autoregressive manner to focusing solely on lab test events has a positive impact on the model's predictive accuracy.
This suggests that selective loss computation allows the model to effectively develop a deeper understanding of patterns and relationships specific to lab test outcomes.
In addition, incorporating a text-based embedding approach further enhances performance.
This improvement again highlights the advantage of providing the model with rich and interpretable representations of medical events, enabling it to better capture relationships within EHR data.

\subsubsection{Numeric Value Representation}
An alternative approach to handling numeric values in language models is discretization into quantiles, where continuous values are mapped to a set of predefined bins and treated as categorical tokens (\textit{i.e.,} \textbf{quantile-based tokenization})~\citep{renc2024zero}.
In contrast, our model treats numeric values as text and tokenizes them at the digit level (\textit{i.e.,} \textbf{digit-wise tokenization}).
To evaluate the effectiveness of this approach against the quantile-based method, we conduct experiments using different levels of quantization (5, 10, and 20 quantiles) and compare the results with the digit-wise tokenization strategy.

For the quantile-based method, the predicted value $\hat{y}$ is computed as the expected value of the quantile probabilities for each lab item.
Specifically, given a set of quantile bins $\mathcal{Q}$, each bin $q\in\mathcal{Q}$ is associated with a probability $p(q)$.
The predicted value is defined as the weighted sum of the expected values of the quantile intervals:
\begin{equation}
    \hat{y}=\sum_{q\in\mathcal{Q}}\mu_q \cdot p(q)
\end{equation}
where $\mu_q$ is the average value of the target lab item samples belonging to the quantile $q$ in the training set.

As shown in Table~\ref{tab:abl_unified}, performance improves consistently as the number of quantile bins increases, suggesting that finer-grained numeric representations lead to better predictions.
Following this trend, digit-wise tokenization, as used in \ours{}, achieves the best performance, likely because it provides the highest resolution representation by preserving the exact numerical value rather than grouping them into fixed bins.
We believe this approach allows the model to capture subtle variations in lab test values that might otherwise be lost in a coarser representation.
Thus, these results highlight the importance of maintaining fine-grained detail when encoding numerical values to process EHR data effectively.

\subsubsection{Timestamp Representation}
As discussed in Section~\ref{sec:abs_time_encoding}, a patient's medical state can be influenced by daily routines and hospital schedules, making absolute time encoding more suitable than relative time encoding.
To validate the effectiveness of our approach, we compare our model trained with \textbf{relative time encoding} against the one trained with \textbf{absolute time encoding}, as presented in Table~\ref{tab:abl_unified}.
For relative time encoding, we calculate temporal intervals in minutes between consecutive events and append each interval to the corresponding medical event $\mathcal{M}_i$ after converting it to a text format, instead of prepending absolute timestamp tokens which we defined.

The result demonstrates that absolute time encoding leads to superior performance compared to relative time encoding, suggesting that absolute time encoding implicitly provides a more meaningful temporal context than merely representing time as intervals between events.
In other words, absolute time encoding allows the model to capture daily patterns and periodic trends that influence lab test results, such as fasting-related fluctuations in glucose levels or routine morning lab draws in ICU settings.
In contrast, relative time encoding only provides information about the time elapsed since the previous event, making it more challenging for the model to infer broader temporal structures.
These findings emphasize the importance of designing time representations that align with the structured nature of clinical workflows.

\begin{figure}[t] 
    \centering
    \includegraphics[width=1\columnwidth]{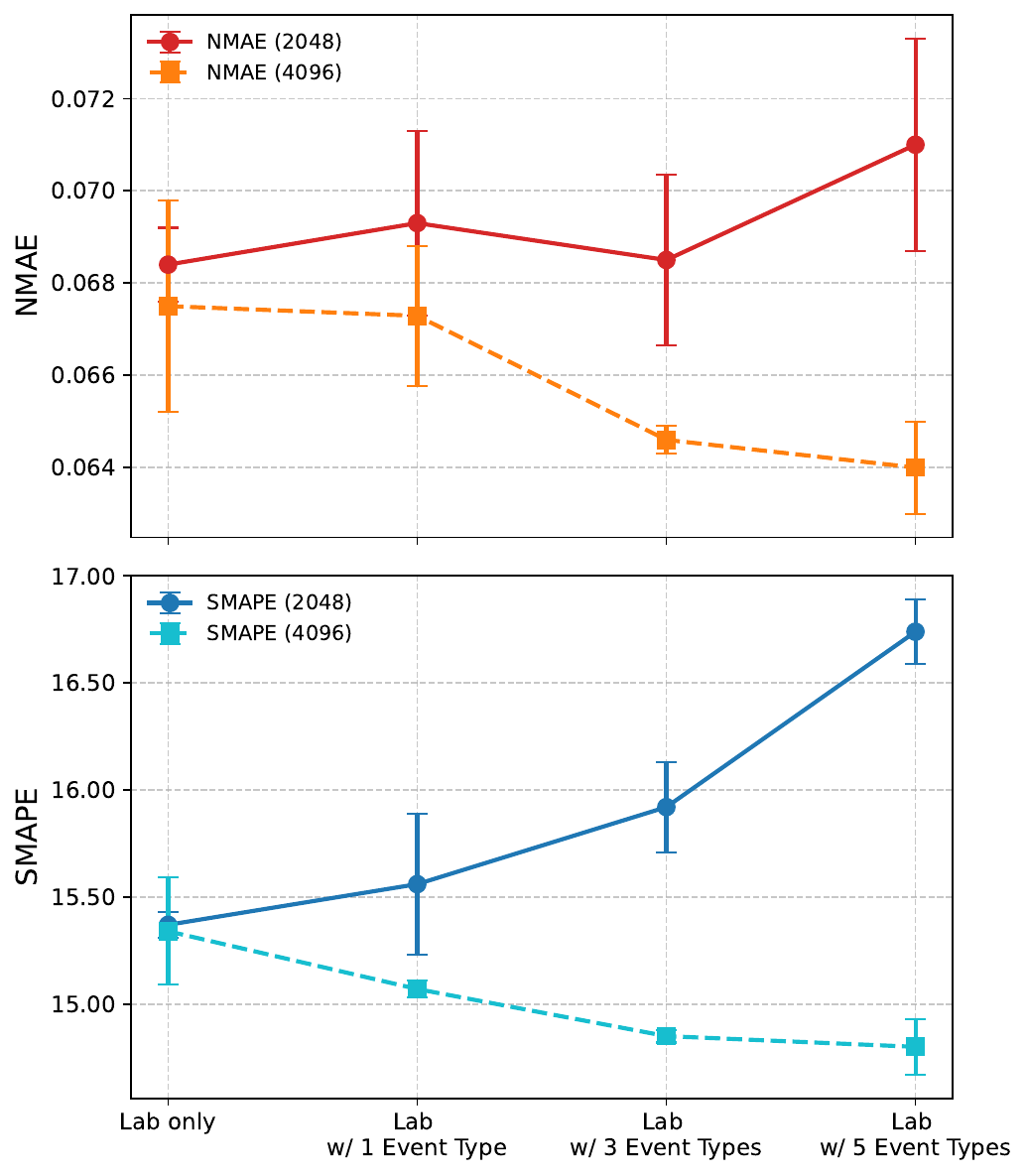} 
    \vspace{-4mm}
    \caption{
    Performance trends across different configurations of event types.
    }
    \label{fig:abl_tables}
    \vspace{-8mm}
\end{figure}

\subsubsection{Combination of Different Event Types}
As described in Section~\ref{sec:data}, we prioritized event types into three levels based on their expected influence on lab test results.
To assess how including a broader range of clinical events affects performances, we train models with different configurations of event types according to the criterion described in Section~\ref{sec:data}, and analyze their performance trends.
Specifically, each configuration incrementally includes (1) medication events (\textit{i.e., emar} and \textit{emar\_detail}), (2) input and procedure events (\textit{i.e., inputevents} and \textit{procedureevents}), and (3) output and microbiology events (\textit{i.e., outputevents} and \textit{microbiologyevents}).
Here, while utilizing more event types can provide richer contextual information, it can also lead the model to see a shorter historical window within the fixed sequence length.
For example, when using only two types of events (lab test and medication), the average covered time span per sample is 28.8 hours with a sequence length of 2048 tokens and 47.2 hours with 4096 tokens.
In contrast, when including six types of events, the average covered time span drops to 18.5 hours for 2048 tokens and 33.3 hours for 4096 tokens.
Hence, to systematically examine this trade-off, we also conduct these experiments by adjusting the sequence length to 2048 tokens from 4096 tokens, and compare the trends between them.
Detailed statistics on the average covered time span for each table combination are provided in Appendix~\ref{app:data_statistics}.

Figure~\ref{fig:abl_tables} illustrates the impact of progressively incorporating more event types on the lab test outcome prediction performances.
The results show that when the sequence length is limited to 2048 tokens, increasing the number of event types leads to a degradation in performance for both NMAE and SMAPE.
This suggests that the reduced temporal coverage outweighs the benefits of incorporating more event types in this constrained setting.
In contrast, when using a sequence length of 4096 tokens, expanding the range of event types improves performance in general, indicating that longer sequences allow the model to leverage additional event information effectively.
These findings suggest that while our current experiments are still constrained to a sequence length of 4096 tokens, further increasing the sequence length may unlock additional benefits from incorporating an even broader range of EHR event types.

\section{Conclusion and Discussion}
In this work, we proposed \textbf{\ours{}}, a unified model for lab test outcome prediction that leverages an autoregressive generative modeling approach on EHR data.
By integrating effective data processing and training strategy, \ours{} has shown the best performance across three public EHR datasets.
Specifically, through evaluations on MIMIC-IV, eICU, and HiRID, LabTOP consistently outperformed both traditional machine learning models and state-of-the-art LLMs, demonstrating its potential for clinical decision support and early detection of critical conditions.
Additionally, our ablation studies highlighted the efficacy of the proposed data processing and training strategy in modeling EHR data for the lab test outcome prediction task.
We believe that \ours{} will serve as an accurate and generalizable framework for lab test outcome prediction, especially in constrained environments where repeated testing is not feasible.

\paragraph{Clinical utility of LabTOP}
Our model allows for the quantification of uncertainty in its predictions, which can be measured using the average entropy of token probabilities when generating lab test values.
For example, if a model outputs ``1 2 3 . 4 5'', we can compute entropies for the generated tokens (``1'', ``2'', ``3'', ``.'', ``4'', and ``5'') and average them to get the uncertainty of the predicted output.
By leveraging this uncertainty estimation, high-confidence predictions can be directly used for clinical decision support, while low-confidence predictions can be flagged for further verification through actual lab testing.
This approach helps prioritize critical tests while reducing unnecessary routine tests.
As a specific scenario, in ICUs and other hospital settings where continuous monitoring is critical, patients frequently undergo multiple daily lab tests to assess their condition.
However, frequent blood draws can lead to increased risk of anemia and additional workload for healthcare providers.
According to \cite{james2018blood}, ICU patients often undergo 1-4 blood tests per day, and repeated phlebotomy is a major contributor to anemia in critically ill patients.
In this scenario, by incorporating LabTOP into clinical workflows, healthcare providers could pre-screen lab test predictions and selectively order confirmatory tests only when predictions exhibit high uncertainty.
This approach would reduce unnecessary testing, helping to minimize patient burden and optimize resource allocation while maintaining diagnostic accuracy.

\paragraph{Limitations and future work}
Our experiments have some limitations as follows.
First, \ours{} requires relatively longer sequence length, leading to increased computational costs during training and inference, especially when we involve more types of EHR events.
To mitigate this, we applied optimization techniques such as mixed-precision training to enhance GPU memory efficiency.
Despite these optimizations, memory usage and inference latency remain key concerns for real-world deployment.
Future work could explore more efficient tokenization strategies to balance precision and computational efficiency.
Second, our experiments were focused on retrospective EHR data, and its real-world applicability in prospective clinical settings remains to be validated.
Further research could assess \ours{}'s performance in real-time hospital environments and evaluate its impact on clinical decision-making.
Finally, generalization beyond ICU settings needs further investigation.
Our current model is designed for short-term (near-future) prediction, leveraging high-density time-series data and trained in an auto-regressive manner.
This setup aligns well with ICU and inpatient settings, where lab tests are recorded at high frequency and recent trends play a crucial role in clinical decision-making.
In contrast, medical records for the general population tend to be recorded at irregular intervals, often spanning months or years between visits.
Under such conditions, our training and inference strategy may not generalize effectively.
Instead of directly applying \ours{} in this setting, a more suitable approach may involve alternative modeling techniques, such as point process-based models or learning strategies specifically designed for sparse and irregular medical data, which would be an interesting future research direction.

\acks{This work was supported by the Institute of Information \& Communications Technology Planning \& Evaluation (IITP) grant (No.RS-2019-II190075), National Research Foundation of Korea (NRF) grant (NRF-2020H1D3A2A03100945, RS-2023-00262527), funded by the Korea government (MSIT).}

\clearpage
\bibliography{chil-sample}

\clearpage

\appendix
\section{Dataset Statistics}
\label{app:data_statistics}
In this appendix, we describe dataset statistics used throughout our experiments. 

\begin{table}[h]
\floatconts
    {tab:app_datastat1}
    {\caption{Number of ICU stays in the training, validation, and test splits for each dataset (MIMIC-IV, eICU, and HiRID).}}
    {
        \begin{tabular}{lccc}
            \toprule
             & MIMIC-IV & eICU & HiRID \\
            \midrule
            Train & 73,236 & 141,712 & 26,597 \\
            Validation & 9,151 & 17,716 & 3,325 \\
            Test & 9,152 & 17,710 & 3,323 \\
            \bottomrule        
        \end{tabular}
    }
\end{table}

\paragraph{Number of ICU records}
Table~\ref{tab:app_datastat1} presents the number of ICU records in the training, validation, and test splits for each dataset (MIMIC-IV, eICU, and HiRID).
We split ICU records based on a patient level, ensuring that records from the same patient do not appear in multiple subsets.
The datasets follow an 8:1:1 split ratio, where 80\% of ICU records are assigned to training, 10\% to validation, and 10\% to test splits.

\begin{table}[h]
\floatconts
    {tab:app_datastat2}
    {\caption{Number of lab events in the training, validation, and test splits for each dataset.}}
    {
        \begin{tabular}{p{1.5cm}ccc}
            \toprule
            & MIMIC-IV & eICU & HiRID \\
            \midrule
            Train & 14,716,753 & 15,225,218 & 3,768,827 \\
            Validation & 1,739,599 & 1,914,090 & 454,994 \\
            Test & 1,799,223 & 1,848,832 & 456,609 \\
            \bottomrule        
        \end{tabular}
    }
\end{table}

\paragraph{Number of lab test events}
Table~\ref{tab:app_datastat2} presents the number of lab test events that appeared in the training, validation, and test splits for each dataset.
Given the substantial volume of data, the model was exposed to a diverse range of lab test events, enabling a thorough evaluation of its ability to predict lab test outcomes.

\begin{table}[t]
\floatconts
    {tab:time_span_mimiciv}
    {\caption{Average time span per sample in hours on MIMIC-IV.}}
    {\resizebox{1.0\linewidth}{!}{
    \begin{tabular}{lcccc}
        \toprule
        Sequence & \multirow{2}{*}{Lab-only} & Lab w/ & Lab w/ & Lab w/ \\
        Length & & T1 & T2 & T3 \\
        \midrule
        2048 & 32.8 & 20.0 & 20.5 & 18.5 \\
        4096 & 43.6 & 34.8 & 23.1 & 20.3 \\
        \midrule
        \multicolumn{5}{l}{\footnotesize T1: \{medication event\}} \\
        \multicolumn{5}{l}{\footnotesize T2: T1 $\cup$ \{input and procedure events\}} \\
        \multicolumn{5}{l}{\footnotesize T3: T2 $\cup$ \{output and microbiology events\}}
    \end{tabular}}
    }
\end{table}

\paragraph{Selected lab items}
To focus on the most clinically relevant medical events, we selected lab tests and other event types based on their frequency, retaining only those that accounted for the top 90\% occurrences within their respective tables.
As a result, we included 44 lab tests from MIMIC-IV, 41 from eICU, and 27 from HiRID, as shown in Table~\ref{tab:lab_items}.
While core biomarkers such as glucose, sodium, potassium, chloride, creatinine, bicarbonate, hemoglobin, platelets, and white blood cell count are present across all datasets, notable differences exist: (1) MIMIC-IV includes a broader set of liver function tests (AST, ALT, bilirubin), coagulation markers (PT, INR), and red blood cell indices (MCV, MCH, RDW); (2) eICU incorporates hematological parameters (MPV, lymphocytes, monocytes) and oxygenation measures (FiO2, O2 saturation, base excess); and (3) HiRID focuses more on arterial blood gas (ABG) parameters (pH, PaO2, PaCO2, lactate, methemoglobin, carboxyhemoglobin), reflecting its emphasis on real-time ICU monitoring.

\paragraph{Time span coverage per sample}
Table~\ref{tab:time_span_mimiciv} presents the average time span per sample (in hours) for different combinations of event types in MIMIC-IV, according to a sequence length of 2048 and 4096 tokens.
The time span represents the total duration covered within each sample's sequence.
As expected, increasing the sequence length from 2048 to 4096 results in a longer time span coverage across all event combinations.
Specifically, for lab-only sequences, the time span increases from 32.8 hours to 43.6 hours.
However, adding more event types reduces the time span per sample, as additional events make each sample more dense within the same sequence length.
For example, when we use a sequence length of 2048 tokens, incorporating one event type (Lab w/ T1) reduces the time span to 20.0 hours, while including two and four more event types (Lab w/ T2 and Lab w/ T3) results in 20.5 hours and 18.5 hours respectively.
Similarly, for a sequence length of 4096 tokens, the time span decreases from 43.6 hours (Lab-only) to 34.8 hours (Lab w/ T1), 23.1 hours (Lab w/ T2), and 20.3 hours (Lab w/ T3).

\begin{table*}[t]
\floatconts
    {tab:lab_items}
    {\caption{Lab items included in each dataset (MIMIC-IV, eICU, and HiRID)}}
    {
        \begin{tabular}{p{4cm}p{10cm}}
            \toprule
            \textbf{Dataset} & \textbf{Lab Items} \\
            \midrule
            MIMIC-IV (44 items) & glucose, potassium, sodium, chloride, ph, hemoglobin, hematocrit, bicarbonate, creatinine, anion gap, urea nitrogen, magnesium, phosphate, calcium, total, platelet count, white blood cells, mchc, red blood cells, mcv, mch, rdw, po2, base excess, calculated total co2, pco2, ptt, pt, inr(pt), h, l, i, lactate, rdw-sd, free calcium, oxygen saturation, potassium, whole blood, asparate aminotransferase (ast), bilirubin, total, alanine aminotransferase (alt), alkaline phosphatase, temperature, lactate dehydrogenase (ld), albumin, lymphocytes \\
            \midrule
            eICU (41 items)& bedside glucose, potassium, sodium, glucose, hgb, chloride, creatinine, bun, hct, calcium, bicarbonate, platelets x 1000, wbc x 1000, rbc, mcv, mchc, mch, rdw, anion gap, magnesium, mpv, -lymphs, -monos, pao2, paco2, ph, hco3, fio2, -eos, phosphate, -polys, -basos, albumin, o2 sat (\%), base excess, ast (sgot), total protein, alt (sgpt), alkaline phos., total bilirubin, pt - inr \\
            \midrule
            HiRID (27 items) & glucose [moles/volume] in serum or plasma, sodium [moles/volume] in blood, potassium [moles/volume] in blood, bicarbonate [moles/volume] in arterial blood, base excess in arterial blood by calculation, oxygen saturation in arterial blood, methemoglobin/hemoglobin.total in arterial blood, carboxyhemoglobin/hemoglobin.total in arterial blood, calcium.ionized [moles/volume] in blood, lactate [mass/volume] in arterial blood, chloride [moles/volume] in blood, ph of arterial blood, hemoglobin [mass/volume] in arterial blood, oxygen [partial pressure] in arterial blood, carbon dioxide [partial pressure] in arterial blood, hemoglobin [mass/volume] in blood, leukocytes [\#/volume] in blood, mcv [entitic volume], mch [entitic mass], mchc [mass/volume] in cord blood, platelets [\#/volume] in blood, inr in blood by coagulation assay, creatinine [moles/volume] in blood, c reactive protein [mass/volume] in serum or plasma, urea [moles/volume] in venous blood, phosphate [moles/volume] in blood, magnesium [moles/volume] in blood
            \\
            \bottomrule        
        \end{tabular}
    }
\end{table*}

\paragraph{Time intervals for each lab item}
Tables~\ref{tab:time_gaps_mimiciv}, \ref{tab:time_gaps_eicu} and \ref{tab:time_gaps_hirid} present the time granularity (in hours) of lab items in MIMIC-IV, eICU, and HiRID, respectively.
The time interval for each lab item represents the average interval (in hours) at which the lab item was recorded.
A comparison of these tables reveals that the intervals between the same lab items are shortest in HiRID, indicating that lab tests are conducted more frequently in HiRID than in the other datasets.

\begin{table*}[h]
\floatconts
    {tab:time_gaps_mimiciv}
    {\caption{Time intervals for each lab item in MIMIC-IV}}
    {
        \begin{tabular}{lc}
            \toprule
           \textbf{Item} & \textbf{Time Interval (Hours)} \\
            \midrule
            Alanine Aminotransferase (ALT) & 25.65 \\
            Albumin & 40.22 \\
            Alkaline Phosphatase & 25.79 \\
            Anion Gap & 13.38 \\
            Aspartate Aminotransferase (AST) & 25.62 \\
            Base Excess & 7.63 \\
            Bicarbonate & 13.34 \\
            Bilirubin, Total & 25.53 \\
            Calcium, Total & 14.09 \\
            Calculated Total CO2 & 7.63 \\
            Chloride & 12.79 \\
            Creatinine & 13.40 \\
            Free Calcium & 10.33 \\
            Glucose & 9.99 \\
            H & 10.94 \\
            Hematocrit & 13.26 \\
            Hemoglobin & 12.80 \\
            I & 10.94 \\
            INR (PT) & 18.41 \\
            L & 10.94 \\
            Lactate & 10.98 \\
            Lactate Dehydrogenase (LD) & 33.55 \\
            Lymphocytes & 47.93 \\
            Magnesium & 13.45 \\
            MCH & 15.05 \\
            MCHC & 15.05 \\
            MCV & 15.05 \\
            Oxygen Saturation & 9.34 \\
            PCO2 & 7.63 \\
            pH & 7.85 \\
            Phosphate & 14.00 \\
            Platelet Count & 14.74 \\
            PO2 & 7.62 \\
            Potassium & 12.43 \\
            Potassium, Whole Blood & 9.93 \\
            PT & 18.41 \\
            PTT & 16.25 \\
            RDW & 15.06 \\
            RDW-SD & 14.41 \\
            Red Blood Cells & 15.05 \\
            Sodium & 12.42 \\
            Temperature & 17.91 \\
            Urea Nitrogen & 13.44 \\
            White Blood Cells & 15.04 \\
            \midrule
            \textbf{Average} & \textbf{15.80} \\
            \bottomrule
        \end{tabular}
    }
\end{table*}
\begin{table*}[t]
\floatconts
    {tab:time_gaps_eicu}
    {\caption{
    Time intervals for each lab item in eICU
    }}
    {
        \begin{tabular}{lc}
            \toprule
            \textbf{Item} & \textbf{Time Interval (Hours)} \\
            \midrule
            Basos & 25.80 \\
            Eos & 25.78 \\
            Lymphs & 24.57 \\
            Monos & 24.60 \\
            Polys & 24.85 \\
            Albumin & 26.75 \\
            Alkaline Phos. & 28.56 \\
            ALT (SGPT) & 28.67 \\
            Anion Gap & 17.54 \\
            AST (SGOT) & 28.62 \\
            Base Excess & 14.68 \\
            Bedside Glucose & 4.10 \\
            Bicarbonate & 17.44 \\
            BUN & 17.57 \\
            Calcium & 17.63 \\
            Chloride & 17.29 \\
            Creatinine & 17.53 \\
            FiO2 & 14.65 \\
            Glucose & 16.64 \\
            HCO3 & 14.13 \\
            Hct & 17.34 \\
            Hgb & 16.82 \\
            Magnesium & 20.06 \\
            MCH & 20.21 \\
            MCHC & 20.19 \\
            MCV & 20.18 \\
            MPV & 20.36 \\
            O2 Sat (\%) & 14.34 \\
            PaCO2 & 13.89 \\
            PaO2 & 13.91 \\
            pH & 13.87 \\
            Phosphate & 21.55 \\
            Platelets × 1000 & 19.97 \\
            Potassium & 14.10 \\
            PT - INR & 24.72 \\
            RBC & 20.24 \\
            RDW & 20.31 \\
            Sodium & 15.43 \\
            Total Bilirubin & 28.46 \\
            Total Protein & 28.62 \\
            WBC × 1000 & 20.24 \\
            \midrule
            \textbf{Average} & \textbf{19.81} \\
            \bottomrule
        \end{tabular}
    }
\end{table*}

\begin{table*}[t]
\floatconts
    {tab:time_gaps_hirid}
    {\caption{
    Time intervals for each lab item in HiRID
    }}
    {
        \begin{tabular}{lc}
            \toprule
            \textbf{Item} & \textbf{Time Interval (Hours)}\\
            \midrule
            Base Excess in Arterial Blood by Calculation & 6.59 \\
            Bicarbonate [moles/volume] in Arterial Blood & 6.59 \\
            C-Reactive Protein [mass/volume] in Serum or Plasma & 24.72 \\
            Calcium, Ionized [moles/volume] in Blood & 6.59 \\
            Carbon Dioxide [partial pressure] in Arterial Blood & 6.41 \\
            Carboxyhemoglobin/Hemoglobin Total in Arterial Blood & 6.58 \\
            Chloride [moles/volume] in Blood & 6.64 \\
            Creatinine [moles/volume] in Blood & 23.74 \\
            Glucose [moles/volume] in Serum or Plasma & 3.85 \\
            Hemoglobin [mass/volume] in Arterial Blood & 6.39 \\
            Hemoglobin [mass/volume] in Blood & 17.67 \\
            INR in Blood by Coagulation Assay & 21.09 \\
            Lactate [mass/volume] in Arterial Blood & 6.68 \\
            Leukocytes [\#/volume] in Blood & 18.26 \\
            Magnesium [moles/volume] in Blood & 26.61 \\
            MCH [entitic mass] & 18.35 \\
            MCHC [mass/volume] in Cord Blood & 18.35 \\
            MCV [entitic volume] & 18.32 \\
            Methemoglobin/Hemoglobin Total in Arterial Blood & 6.57 \\
            Oxygen [partial pressure] in Arterial Blood & 6.39 \\
            Oxygen Saturation in Arterial Blood & 6.57 \\
            pH of Arterial Blood & 6.31 \\
            Phosphate [moles/volume] in Blood & 24.72 \\
            Platelets [\#/volume] in Blood & 18.74 \\
            Potassium [moles/volume] in Blood & 6.82 \\
            Sodium [moles/volume] in Blood & 6.77 \\
            Urea [moles/volume] in Venous Blood & 24.43 \\
            \midrule
            \textbf{Average} & \textbf{12.99} \\
            \bottomrule
        \end{tabular}
    }
\end{table*}

\section{Implementation Details}
\label{app:implementation_details}

This appendix provides implementation details, including data processing methods, model configurations, training hyperparameters, data processing steps, and hardware specifications used in the experiments.

\paragraph{\ours{}}
The model architecture of LabTOP follows GPT-2, specifically implemented using the architecture of HuggingFace GPT2LMHeadModel\footnote{\url{https://huggingface.co/docs/transformers/model_doc/gpt2##transformers.GPT2LMHeadModel}}.
We use 12 attention decoder layers, each with 8 heads and a hidden dimension of 512.
All the experiments in this study are conducted using a maximum sequence length of 4096 tokens by default.
If a sequence of an ICU record exceeds the maximum sequence length, we repeatedly crop the sequence by this maximum length based on the event level.
Additionally, we again prepend demographic information $D$ to all the cropped samples.
For training, we applied early stopping with the patience of 5 epochs based on the validation loss, which yielded about $120$k, $110$k, and $34$k training steps with the learning rate of 1e-4 for MIMIC-IV, eICU, and HiRID, respectively.
As a result, we trained the model for approximately 96, 92, and 72 hours with 2 A6000 48GB GPUs for each dataset, respectively.

\paragraph{Naive}
If no prior measurement exists for the target lab test within the corresponding ICU record, the prediction is set to the mean of that lab test's values computed across the training set.

\paragraph{Naive($\mu$)}
If no prior measurements exist, the prediction defaults to the mean of that lab test's values across the training set, similar to the Naive method.

\paragraph{GenHPF}
GenHPF has a hierarchical architecture consisting of an event encoder and an event aggregator, designed to perform multiple predictive tasks in a single framework.
To adapt it for regressing multiple lab test outcomes, we construct samples by grouping the original samples that have the same prior medical history (\textit{i.e.,} lab test samples that occurred at the same timestamp).
Then, we manipulate the GenHPF model to regress every single lab test outcome given a sequence of medical events but compute the Mean Squared Error (MSE) loss only for the observed lab items for which we can define the ground-truth labels.
Here, since lab items with large magnitudes may disproportionately influence the model training, we standardize the outcome values for each lab item using the mean and standard deviation computed from the training set.
We maintain GenHPF's core architecture, with key hyperparameters including a prediction dimension of 128, an embedding dimension of 128, 4 attention heads, and 2 transformer layers.
Similar to LabTOP, we applied early stopping with the patience of 5 epochs based on the validation loss, which resulted in about $83$k, $210$k, and $71$k training steps with the learning rate of 3e-4 for each dataset, respectively.
As a result, we trained the model for approximately $39$, $67$, and $30$ hours with a single 80GB A100 GPU for each respective dataset.


\paragraph{XGBoost}
For XGBoost, the input features include the count, mean, min, and max values for all medical items used in the dataset, which are treated as separate columns.
As a result, the input consists of an array of the statistical summaries (count, mean, min, and max) of the unique events that precede the prediction time for the target lab item.
We trained this model for each lab item existed in the dataset with a learning rate of 0.1, a maximum depth of 5, and 100 estimators.

\begin{figure*}[t] 
    \centering
    \includegraphics[width=1.0\textwidth]{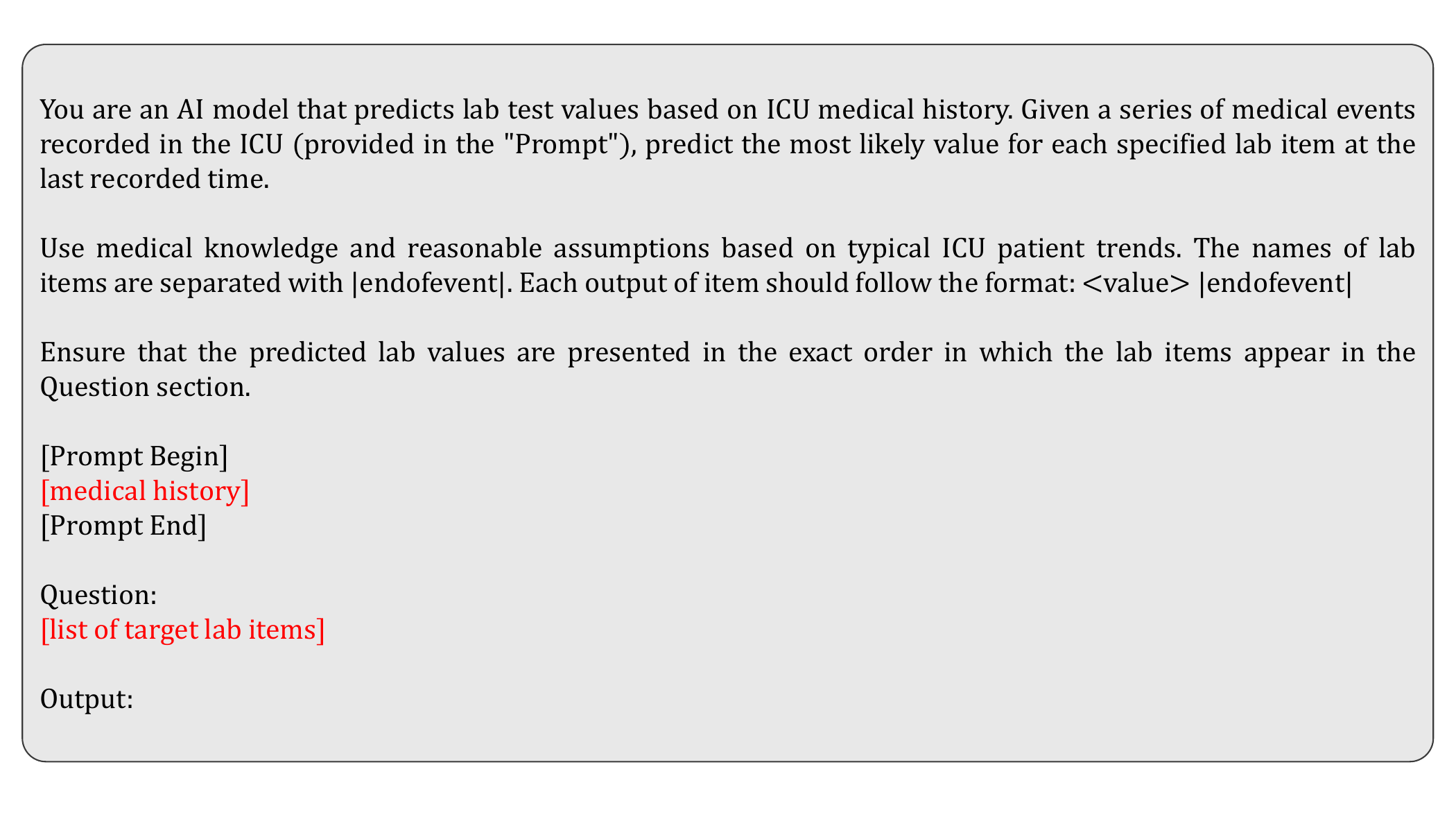} 
    \vspace{-4mm}
    \caption{
    Template prompt used for GPT-4o and GPT-4o-mini.
    Note that [medical history] refers to the textual sequence of the test sample.
    }
    \label{fig:prompt}
\end{figure*}

\paragraph{GPT-4o, GPT-4o-mini, and LLaMA-3.1-Instruct (8B)}
To create samples for these LLMs\footnote{Specifically, at the time of the experiments, GPT-4o referred to \texttt{gpt-4o-2024-11-20} and GPT-4o-mini corresponded to \texttt{gpt-4o-mini-2024-07-18}.}, we followed the same strategy with GenHPF.
In other words, we grouped test samples that have the same prior medical history (\textit{i.e.,} the lab items that occurred at the same time), and the LLM was instructed to answer values of the target lab items based on their prior sequence of medical events as prompt.
The prompt we used for lab test outcome prediction can be found in Figure~\ref{fig:prompt}.

\begin{table*}[h]
\floatconts
    {tab:mae_mimiciv}
    {\caption{LabTOP performance for each individual lab item in MIMIC-IV}}
    {
    \begin{tabular}{lclll}
    \toprule
                                 Item &  Count &                  MAE &                SMAPE &    Unit \\
    \midrule
                              Glucose &  72887 &  $48.81_{\pm{2.39}}$ &  $20.04_{\pm{0.44}}$ &   mg/dL \\
                               Sodium &  60140 &    $1.40_{\pm{1.09}}$ &   $3.27_{\pm{3.64}}$ &   mEq/L \\
                            Potassium &  60131 &   $0.84_{\pm{1.08}}$ &   $5.12_{\pm{3.64}}$ &   mEq/L \\
                                   Ph &  59069 &    $0.20_{\pm{0.06}}$ &   $2.69_{\pm{0.18}}$ &   units \\
                             Chloride &  59000 &   $2.12_{\pm{0.01}}$ &   $2.06_{\pm{0.01}}$ &   mEq/L \\
                           Hemoglobin &  57786 &    $0.65_{\pm{0.00}}$ &   $6.72_{\pm{0.03}}$ &    g/dL \\
                          Bicarbonate &  56749 &    $1.61_{\pm{0.00}}$ &    $6.83_{\pm{0.00}}$ &   mEq/L \\
                           Hematocrit &  56497 &    $1.88_{\pm{0.00}}$ &   $6.33_{\pm{0.01}}$ &      \% \\
                            Anion Gap &  56445 &   $1.79_{\pm{0.01}}$ &  $13.18_{\pm{0.06}}$ &   mEq/L \\
                        Urea Nitrogen &  56406 &   $4.32_{\pm{0.01}}$ &  $15.84_{\pm{0.03}}$ &   mg/dL \\
                           Creatinine &  56086 &   $0.22_{\pm{0.04}}$ &  $13.62_{\pm{0.06}}$ &   mg/dL \\
                            Magnesium &  54857 &   $0.13_{\pm{0.01}}$ &   $6.29_{\pm{0.01}}$ &   mg/dL \\
                            Phosphate &  52445 &    $0.51_{\pm{0.00}}$ &  $14.71_{\pm{0.04}}$ &   mg/dL \\
                       Calcium, Total &  51933 &   $0.32_{\pm{0.03}}$ &   $3.57_{\pm{0.02}}$ &   mg/dL \\
                       Platelet Count &  51197 &  $29.21_{\pm{0.21}}$ &  $16.73_{\pm{0.15}}$ &    K/uL \\
                                  Po2 &  50431 &   $31.6_{\pm{0.16}}$ &  $26.81_{\pm{0.25}}$ &   mm Hg \\
                          Base Excess &  50431 &    $1.33_{\pm{0.00}}$ &  $65.21_{\pm{0.51}}$ &   mEq/L \\
                                 Pco2 &  50379 &   $4.23_{\pm{0.01}}$ &   $9.76_{\pm{0.03}}$ &   mm Hg \\
                 Calculated Total Co2 &  50329 &   $1.42_{\pm{0.01}}$ &   $5.72_{\pm{0.01}}$ &   mEq/L \\
                    White Blood Cells &  50194 &   $2.27_{\pm{0.01}}$ &  $19.75_{\pm{0.22}}$ &    K/uL \\
                      Red Blood Cells &  50153 &    $0.22_{\pm{0.0}}$ &   $6.71_{\pm{0.01}}$ &    m/uL \\
                                  Mch &  50122 &   $0.65_{\pm{0.01}}$ &   $2.19_{\pm{0.01}}$ &     pg \\
                                  Rdw &  50109 &   $0.51_{\pm{0.01}}$ &   $3.22_{\pm{0.04}}$ &       \% \\
                                 Mchc &  50066 &    $0.64_{\pm{0.0}}$ &   $1.96_{\pm{0.01}}$ & \%, g/dL \\
                                  Mcv &  50045 &    $1.70_{\pm{0.01}}$ &   $1.85_{\pm{0.01}}$ &    g/dL \\
                                  Ptt &  37743 &   $8.95_{\pm{0.24}}$ &  $15.43_{\pm{0.36}}$ &     sec \\
                           Inr ( Pt ) &  34670 &   $0.18_{\pm{0.01}}$ &  $10.36_{\pm{0.14}}$ &    None \\
                                   Pt &  34659 &   $1.98_{\pm{0.02}}$ &   $10.14_{\pm{0.10}}$ &     sec \\
                                    L &  32700 &   $5.69_{\pm{0.02}}$ &  $37.54_{\pm{0.14}}$ &       U \\
                              Lactate &  31327 &    $0.55_{\pm{0.00}}$ &  $23.28_{\pm{0.28}}$ &  mmol/L \\
                                    H &  30831 &  $11.41_{\pm{0.31}}$ & $104.29_{\pm{7.62}}$ &       U \\
                             Rdw - Sd &  29442 &   $1.93_{\pm{0.02}}$ &   $3.61_{\pm{0.04}}$ &      fL \\
                         Free Calcium &  28758 &    $0.05_{\pm{0.0}}$ &   $4.27_{\pm{0.07}}$ &  mmol/L \\
                                    I &  25471 &    $0.32_{\pm{0.0}}$ &  $17.09_{\pm{0.22}}$ &       U \\
                    Oxygen Saturation &  18239 &   $9.31_{\pm{0.16}}$ &   $12.37_{\pm{0.30}}$ &      \% \\
               Potassium, Whole Blood &  16293 &   $0.32_{\pm{0.04}}$ &   $7.08_{\pm{0.01}}$ &   mEq/L \\
    Asparate Aminotransferase ( Ast ) &  14980 &  $81.23_{\pm{0.79}}$ &  $43.65_{\pm{0.39}}$ &    IU/L \\
                 Alkaline Phosphatase &  14696 &  $31.38_{\pm{0.18}}$ &  $23.61_{\pm{0.12}}$ &    IU/L \\
     Alanine Aminotransferase ( Alt ) &  14582 &  $53.34_{\pm{0.34}}$ &  $44.77_{\pm{0.68}}$ &    IU/L \\
                     Bilirubin, Total &  13902 &    $0.70_{\pm{0.01}}$ &  $34.28_{\pm{0.23}}$ &   mg/dL \\
                          Temperature &  10331 &   $0.47_{\pm{0.02}}$ &   $1.29_{\pm{0.02}}$ &      °C \\
         Lactate Dehydrogenase ( Ld ) &   8792 & $177.57_{\pm{2.22}}$ &  $29.48_{\pm{0.21}}$ &    IU/L \\
                              Albumin &   7552 &   $0.31_{\pm{0.01}}$ &  $10.69_{\pm{0.11}}$ &    g/dL \\
                          Lymphocytes &   7488 &    $7.00_{\pm{0.18}}$ &  $70.41_{\pm{3.78}}$ &      \% \\
    \bottomrule
    \end{tabular}
    }
\end{table*}

\begin{table*}[t]
\floatconts
    {tab:mae_eicu}
    {\caption{LabTOP performance for each individual lab item in eICU}}
    {
        \begin{tabular}{lclll}
        \toprule
                    Item &  Count &                 MAE &                SMAPE &    Unit \\
        \midrule
         Bedside Glucose & 179312 & $22.83_{\pm{1.78}}$ &  $15.16_{\pm{0.03}}$ &   mg/dL \\
               Potassium &  77648 &   $0.32_{\pm{0.00}}$ &   $8.08_{\pm{0.02}}$ &  mmol/L \\
                  Sodium &  71184 &  $2.22_{\pm{0.01}}$ &   $1.62_{\pm{0.03}}$ &  mmol/L \\
                 Glucose &  67292 & $22.53_{\pm{0.02}}$ &  $15.58_{\pm{0.01}}$ &   mg/dL \\
                     Hgb &  65241 &  $0.84_{\pm{0.03}}$ &   $8.03_{\pm{0.02}}$ &    g/dL \\
                Chloride &  64786 &  $2.53_{\pm{0.01}}$ &   $2.44_{\pm{0.01}}$ &  mmol/L \\
              Creatinine &  64094 &   $0.29_{\pm{0.00}}$ &   $18.90_{\pm{0.02}}$ &   mg/dL \\
                     Hct &  63676 &  $2.43_{\pm{0.01}}$ &   $7.95_{\pm{0.04}}$ &       \% \\
                     Bun &  63539 &  $5.98_{\pm{0.02}}$ &  $22.59_{\pm{0.03}}$ &   mg/dL \\
                 Calcium &  61862 &   $0.33_{\pm{0.00}}$ &   $4.04_{\pm{0.01}}$ &   mg/dL \\
             Bicarbonate &  61027 &  $1.94_{\pm{0.01}}$ &    $8.10_{\pm{0.01}}$ &  mmol/L \\
        Platelets X 1000 &  56096 &  $34.39_{\pm{0.1}}$ &  $19.33_{\pm{0.02}}$ &   K/mcL \\
              Wbc X 1000 &  55922 &   $2.70_{\pm{0.01}}$ &  $23.17_{\pm{0.08}}$ &   K/mcL \\
                     Rbc &  55748 &  $0.29_{\pm{0.01}}$ &   $8.19_{\pm{0.01}}$ &   M/mcL \\
                     Mcv &  54438 &  $2.16_{\pm{0.33}}$ &   $2.23_{\pm{0.01}}$ &      fL \\
                    Mchc &  54131 &   $0.63_{\pm{0.00}}$ &   $1.93_{\pm{0.01}}$ &    g/dL \\
                     Mch &  51772 &  $0.75_{\pm{0.01}}$ &   $2.55_{\pm{0.01}}$ &      pg \\
                     Rdw &  51522 &  $0.68_{\pm{0.01}}$ &   $4.27_{\pm{0.02}}$ &       \% \\
               Anion Gap &  50748 &  $1.98_{\pm{0.01}}$ &  $20.21_{\pm{0.02}}$ &     nan \\
               Magnesium &  38416 &   $0.20_{\pm{0.03}}$ &   $9.61_{\pm{0.01}}$ &   mg/dL \\
                     Mpv &  37782 &   $0.48_{\pm{0.00}}$ &    $4.90_{\pm{0.01}}$ &      fL \\
                    Pao2 &  29405 &  $35.90_{\pm{0.14}}$ &   $29.50_{\pm{0.05}}$ &   mm Hg \\
                   Paco2 &  29181 &  $5.46_{\pm{0.01}}$ &  $12.75_{\pm{0.02}}$ &   mm Hg \\
                      Ph &  28965 &  $0.07_{\pm{0.03}}$ &   $0.67_{\pm{0.03}}$ &     nan \\
                   Monos &  28891 &  $2.69_{\pm{0.01}}$ &  $46.81_{\pm{0.06}}$ &       \% \\
                  Lymphs &  28774 &  $4.64_{\pm{0.14}}$ &  $46.58_{\pm{0.01}}$ &       \% \\
                    Hco3 &  27738 &  $1.89_{\pm{0.01}}$ &   $8.18_{\pm{0.02}}$ &  mmol/L \\
               Phosphate &  26150 &   $0.67_{\pm{0.00}}$ &  $20.53_{\pm{0.03}}$ &   mg/dL \\
                   Polys &  25945 &  $6.63_{\pm{0.01}}$ &   $9.43_{\pm{0.03}}$ &       \% \\
                    Fio2 &  23978 & $10.51_{\pm{0.04}}$ &   $24.87_{\pm{0.10}}$ &       \% \\
                 Albumin &  23545 &   $0.28_{\pm{0.00}}$ &  $10.58_{\pm{0.03}}$ &    g/dL \\
             Base Excess &  23264 &  $2.37_{\pm{0.35}}$ &  $75.72_{\pm{0.11}}$ &   mEq/L \\
           O2 Sat ( \% ) &  21639 &  $2.89_{\pm{0.01}}$ &   $3.14_{\pm{0.01}}$ &       \% \\
            Ast ( Sgot ) &  20840 & $86.46_{\pm{0.95}}$ &  $52.92_{\pm{0.13}}$ & Units/L \\
           Total Protein &  20677 &  $0.43_{\pm{0.01}}$ &   $7.48_{\pm{0.01}}$ &    g/dL \\
            Alt ( Sgpt ) &  20529 &  $54.60_{\pm{0.98}}$ &  $46.49_{\pm{0.29}}$ & Units/L \\
          Alkaline Phos. &  20391 & $25.44_{\pm{0.14}}$ &  $23.59_{\pm{0.03}}$ & Units/L \\
         Total Bilirubin &  19399 &  $0.53_{\pm{0.01}}$ &  $38.07_{\pm{0.13}}$ &   mg/dL \\
                Pt - Inr &  18298 &   $0.30_{\pm{0.01}}$ &  $16.27_{\pm{0.08}}$ &   ratio \\
                     Eos &  17160 &  $1.22_{\pm{0.01}}$ & $107.39_{\pm{0.21}}$ &       \% \\
                   Basos &  10767 &   $0.31_{\pm{0.00}}$ &  $92.78_{\pm{0.17}}$ &       \% \\
        \bottomrule
        \end{tabular}
    }
\end{table*}

\begin{table*}[h]
\floatconts
    {tab:mae_hirid}
    {\caption{LabTOP performance for each individual lab item in HiRID}}
    {
       \begin{tabular}{lclll}
\toprule
                                              Item &  Count &                 MAE &               SMAPE &   Unit \\
\midrule
     Glucose [ Moles / Volume ] In Serum Or Plasma &  45573 &  $1.28_{\pm{0.01}}$ &  $15.40_{\pm{0.05}}$ & mmol/L \\
                Sodium [ Moles / Volume ] In Blood &  24856 &  $1.72_{\pm{0.01}}$ &  $1.26_{\pm{0.01}}$ & mmol/L \\
             Potassium [ Moles / Volume ] In Blood &  24587 &   $0.26_{\pm{0.00}}$ &  $6.27_{\pm{0.02}}$ & mmol/L \\
  Bicarbonate [ Moles / Volume ] In Arterial Blood &  22615 &  $1.32_{\pm{0.01}}$ &  $5.69_{\pm{0.05}}$ & mmol/L \\
      Base Excess In Arterial Blood By Calculation &  22582 &  $1.46_{\pm{0.01}}$ & $74.43_{\pm{0.92}}$ & mmol/L \\
      Calcium. Ionized [ Moles / Volume ] In Blood &  22396 &   $0.03_{\pm{0.00}}$ &  $2.92_{\pm{0.01}}$ & mmol/L \\
Carboxyhemoglobin / Hemoglobin. Total In Arterial Blood &  22389 &   $0.27_{\pm{0.00}}$ & $21.14_{\pm{0.07}}$ &      \% \\
Methemoglobin / Hemoglobin. Total In Arterial Blood &  22346 &   $0.27_{\pm{0.00}}$ & $31.81_{\pm{0.21}}$ &      \% \\
       Lactate [ Mass / Volume ] In Arterial Blood &  22321 &    $0.50_{\pm{0.00}}$ & $26.95_{\pm{0.17}}$ & mmol/L \\
              Chloride [ Moles / Volume ] In Blood &  21994 &  $1.88_{\pm{0.01}}$ &  $1.73_{\pm{0.02}}$ & mmol/L \\
               Oxygen Saturation In Arterial Blood &  21781 &  $1.57_{\pm{0.05}}$ &  $1.59_{\pm{0.02}}$ &      \% \\
    Hemoglobin [ Mass / Volume ] In Arterial Blood &  20697 &  $6.19_{\pm{0.05}}$ &   $6.10_{\pm{0.09}}$ &    g/L \\
                              Ph Of Arterial Blood &  20673 &   $0.03_{\pm{0.00}}$ &  $0.43_{\pm{0.01}}$ &   None \\
     Oxygen [ Partial Pressure ] In Arterial Blood &  20586 &  $22.02_{\pm{0.30}}$ & $20.11_{\pm{0.11}}$ &   mmHg \\
Carbon Dioxide [ Partial Pressure ] In Arterial Blood &  20313 &  $3.38_{\pm{0.01}}$ &   $9.20_{\pm{0.04}}$ &   mmHg \\
             Hemoglobin [ Mass / Volume ] In Blood &  10373 &  $6.02_{\pm{0.08}}$ &  $5.94_{\pm{0.07}}$ &    g/L \\
                Leukocytes [ \# / Volume ] In Blood &  10124 &  $3.12_{\pm{0.04}}$ & $28.34_{\pm{0.09}}$ &    G/L \\
              Mchc [ Mass / Volume ] In Cord Blood &  10037 &  $6.59_{\pm{0.09}}$ &  $1.95_{\pm{0.02}}$ &    g/L \\
                            Mcv [ Entitic Volume ] &  10016 &  $2.93_{\pm{0.04}}$ &  $3.19_{\pm{0.01}}$ &     fL \\
                              Mch [ Entitic Mass ] &   9992 &   $1.12_{\pm{0.00}}$ &  $3.62_{\pm{0.02}}$ &     pg \\
                 Platelets [ \# / Volume ] In Blood &   9794 & $59.81_{\pm{0.63}}$ & $39.99_{\pm{0.11}}$ &    G/L \\
            Creatinine [ Moles / Volume ] In Blood &   7424 &  $35.31_{\pm{1.20}}$ & $29.87_{\pm{0.35}}$ & umol/L \\
C Reactive Protein [ Mass / Volume ] In Serum or Plasma &   6834 &   $66.40_{\pm{1.20}}$ & $84.87_{\pm{0.53}}$ &   mg/L \\
           Urea [ Moles / Volume ] In Venous Blood &   6248 &  $3.64_{\pm{0.06}}$ & $35.26_{\pm{0.28}}$ & mmol/L \\
                 Inr In Blood By Coagulation Assay &   6010 &   $0.16_{\pm{0.00}}$ &  $12.10_{\pm{0.41}}$ &   None \\
             Phosphate [ Moles / Volume ] In Blood &   5385 &   $0.26_{\pm{0.00}}$ & $23.23_{\pm{0.24}}$ & mmol/L \\
             Magnesium [ Moles / Volume ] In Blood &   5195 &   $0.12_{\pm{0.00}}$ & $12.85_{\pm{0.02}}$ & mmol/L \\
\bottomrule
\end{tabular}
    }
\end{table*}

\section{Additional Experimental Results}

\paragraph{Performances by the time interval to the last measured time for the target lab test}
To further illustrate the performance of our models and Naive model from another perspective, we conduct an additional analysis by grouping test set samples based on the time interval between the last measured time and the prediction time for the target lab item.
Specifically, we categorized samples by whether the time interval to the last measured time for the target lab item is within 24 hours or not, and measured performances within each group. The results are shown in Table~\ref{tab:time_gaps_naive_labtop_mimiciv}.

We observe that when the time interval between the last lab measurement and the prediction time is more than 24 hours, the performance gap between LabTOP and Naive becomes more significant.
Since the Naive model directly copies the last observed value as the prediction, its performance remains relatively strong when lab values are measured frequently and do not change dynamically over short time spans.
However, in longer time interval cases, where patient conditions may have changed significantly, the Naive model struggles while LabTOP maintains higher predictive accuracy (SMAPE).
This result suggests that more computationally advanced methods like LabTOP are particularly valuable in clinically meaningful scenarios, where lab tests are infrequent and accurate forecasting is essential for early detection and intervention.
Therefore, while the computational cost of LabTOP is higher, this trade-off is necessary to achieve reliable performance in challenging clinical cases.

\begin{table*}[h]
\floatconts
    {tab:time_gaps_naive_labtop_mimiciv}
    {\caption{Lab test outcome prediction performances of LabTOP and Naive on MIMIC-IV, aggregated by whether the interval to the last measured time for the target lab test is within 24 hours or not}}
    {
    \begin{tabular}{lcc}
        \toprule
         & $<$24h & $\geq$ 24h \\
        \midrule
        \multicolumn{3}{l}{\textit{\textbf{NMAE}}} \\
        Naive & 0.072 & 0.076 \\
        LabTOP & $0.054_{\pm{0.000}}$ & $0.051_{\pm{0.000}}$ \\
        \midrule
        \multicolumn{3}{l}{\textit{\textbf{SMAPE (\%)}}} \\
        Naive & 15.91 & 20.39 \\
        LabTOP & $12.46_{\pm{0.23}}$ & $12.44_{\pm{0.21}}$ \\
        \bottomrule
    \end{tabular}
}
\end{table*}

\paragraph{LabTOP performances for individual lab items}
The experimental results of LabTOP for each individual lab item in MIMIC-IV, eICU, and HiRID are presented in Table~\ref{tab:mae_mimiciv}, Table~\ref{tab:mae_eicu}, and Table~\ref{tab:mae_hirid}, respectively.

\begin{table*}[t]
\floatconts
    {tab:app_normal_abnormal}
    {\caption{
    Lab test outcome prediction performances for normal-range and abnormal-range samples in MIMIC-IV.
    The best performances for each category (Normal, Abnormal) are highlighted with \textbf{boldface}.
    }}
    {
        \begin{tabular}{lcc}
            \toprule
             & Normal & Abnormal \\
            \midrule
            \multicolumn{3}{l}{\textit{\textbf{NMAE}}} \\
            Naive & $0.081$ & $0.104$ \\
            Naive($\mu$) & $0.093$ & $0.131$ \\
            GenHPF & $0.099_{\pm 0.015}$ & $0.132_{\pm 0.012}$ \\
            XGBoost & $0.070$ & $0.105$ \\
            GPT-4o & $0.196$ & $0.209$ \\
            GPT-4o-mini & $0.805$ & $0.738$ \\
            LLaMA-3.1-Inst. & $1.340$ & $1.115$ \\
            \noalign{\vskip 0.4ex}
            \hline
            \noalign{\vskip 0.4ex}
            \textbf{LabTOP} & $\bm{0.051_{\pm 0.001}}$ & $\bm{0.084_{\pm 0.001}}$ \\
            \midrule
            \multicolumn{3}{l}{\textit{\textbf{SMAPE (\%)}}} \\
            Naive & $16.94$ & $17.80$ \\
            Naive($\mu$) & $19.80$ & $19.30$ \\
            GenHPF & $26.42_{\pm 0.92}$ & $26.52_{\pm 3.08}$ \\
            XGBoost & $22.61$ & $23.45$ \\
            GPT-4o & $20.81$ & $22.92$ \\
            GPT-4o-mini & $41.90$ & $46.11$ \\
            LLaMA-3.1-Inst. & $71.74$ & $79.64$ \\
            \midrule
            \textbf{LabTOP} & $\bm{15.25_{\pm 0.41}}$ & $\bm{17.58_{\pm 0.16}}$ \\
            \bottomrule
        \end{tabular}
    }
\end{table*}

\paragraph{Model performances for normal and abnormal range values in MIMIC-IV}
We group each lab test sample in the test split into normal and abnormal ranges using reference ranges provided in MIMIC-IV, and evaluate models separately for each group.
These results are provided in Table~\ref{tab:app_normal_abnormal}.

\paragraph{Exploration of different model configurations}
We evaluate LabTOP's performance at different model scales by varying the number of Transformer decoder layers, attention heads, and hidden dimension size.
Specifically, we experiment with configurations comparable to BERT-mini, BERT-small, and BERT-base to analyze the effect of model size.
The results of these experiments are presented in Table~\ref{tab:model_size}.

\begin{table*}[t]
\floatconts
    {tab:model_size}
    {\caption{Lab test outcome prediction performances of LabTOP with different model configurations. Note that LabTOP-base is used by default for all the experiments in this paper.}}
    {
        \begin{tabular}{lcccc|cc}
        \toprule
        Model & Layers & Attention Heads & Hidden Dimension & Parameter Size & NMAE & SMAPE \\
        \midrule
        LabTOP-mini     & 4  & 4  & 256  & 11M  & $0.069_{\pm 0.003}$ & $16.02_{\pm 0.53}$  \\
        LabTOP-small    & 4  & 8  & 512  & 30M  & $0.067_{\pm 0.005}$ & $16.24_{\pm 1.28}$ \\
        LabTOP-base     & 12 & 8  & 512  & 53M  & $0.064_{\pm 0.001}$ & $14.80_{\pm 0.13}$ \\
        LabTOP-large    & 12 & 12 & 768  & 110M & $0.063_{\pm 0.003}$ & $15.06_{\pm 0.62}$ \\
        \bottomrule
    \end{tabular}
    }
\end{table*}

\begin{figure*}[t]
    \centering
    \includegraphics[width=2.0\columnwidth,height=0.7\paperheight]{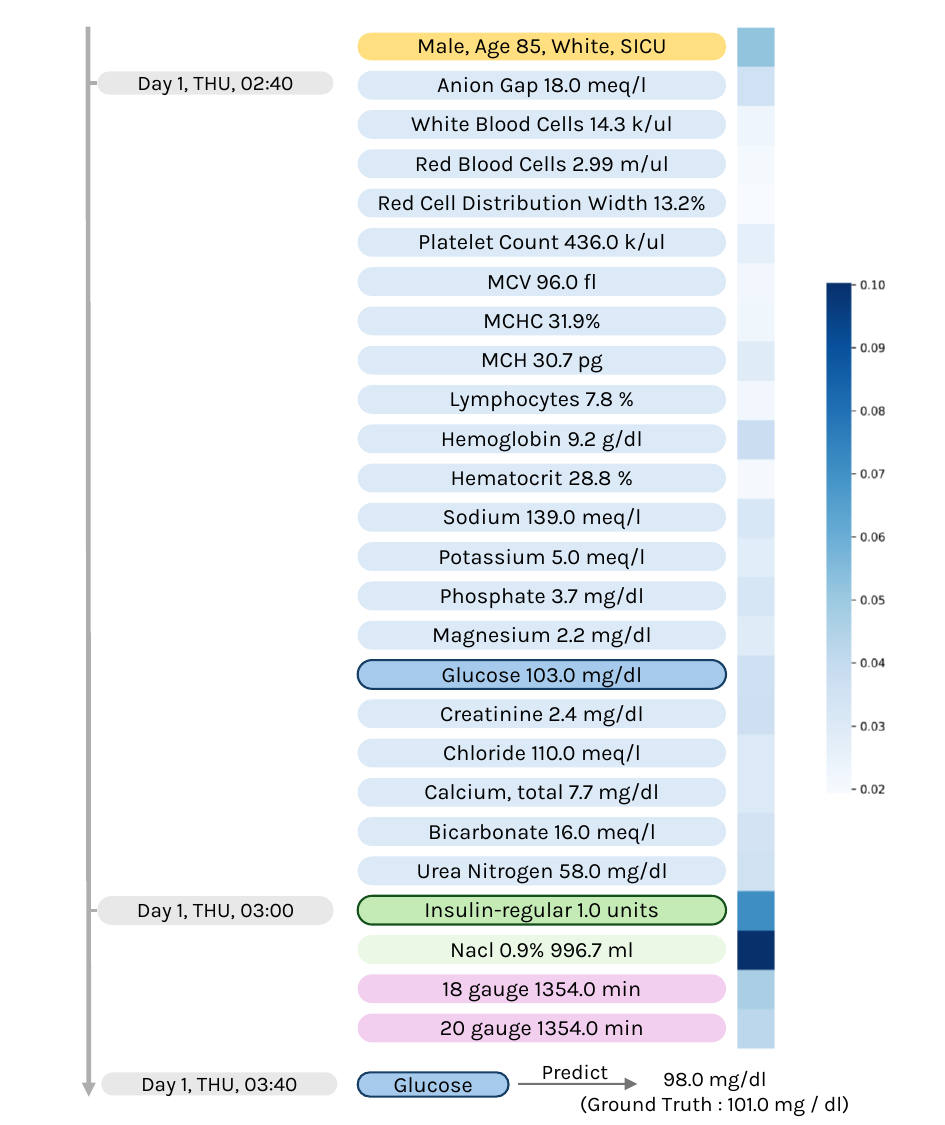} 
    \caption{
    An example of an attention map for LabTOP computed when predicting the glucose outcome
    }
    \label{fig:attn_map_final}
\end{figure*}

\paragraph{Attention Map Analysis}
To gain deeper insights into the prediction process of LabTOP, we can analyze an attention map to see how the model weighs different past clinical events when predicting a specific lab value.
Here, since attention scores are generated for each predicted token across multiple layers and heads, these scores are systematically aggregated to construct a comprehensive attention representation for lab test outcome prediction.
For each layer, the mean attention scores across all heads are computed, and these head-level averaged scores are then used to calculate the mean attention score for each layer.
The aggregated attention score at each layer level is defined as:

\begin{equation}
A_{\text{final}} = \frac{1}{L} \sum_{l=1}^{L} \left( \frac{1}{H} \sum_{h=1}^{H} A_h^{(l)} \right)
\end{equation}

where $A_{\text{final}}$ represents the final aggregated attention score, $L$ is the total number of layers, $H$ is the total number of attention heads, and $A_h^{(l)}$ denotes the attention score of head $h$ in layer $l$.
By averaging the attention maps computed for each predicted token, the final aggregated attention score is obtained.
This resulting attention map captures attention scores from all preceding tokens before the target lab event.

To analyze attention at the event level, tokens before the target lab event are grouped based on their corresponding events.
The total attention score for each event is computed by summing the attention scores of all tokens associated with that event
Event-level attention scores are then visualized to determine which past events are most influential in predicting the lab test outcome.

Figure~\ref{fig:attn_map_final} illustrates the aggregated attention map during the prediction of a glucose outcome.
The glucose level in the preceding events is recorded as $103.0\ mg/dL$.
The model assigned a relatively high attention score to insulin-regular, which is the key factor in predicting a subsequent decrease in glucose levels.
Since both the model prediction and actual glucose measurement confirmed this decrease, the analysis validates that the model effectively leverages relevant prior events to generate accurate predictions.

\end{document}